\documentclass[preprint,12pt]{elsarticle}

\usepackage{float}
\usepackage[margin=1in]{geometry}
\usepackage{amsmath}
\usepackage{booktabs}
\usepackage{tabularx}

\usepackage{graphicx} 
\usepackage{caption} 
\usepackage{subcaption}

\usepackage{amssymb}
\usepackage{color}

\usepackage[table]{xcolor}
\usepackage{adjustbox} 
\usepackage{multirow} 
\usepackage{colortbl}
\usepackage{hyperref}
\hypersetup{
    colorlinks=false,        
    linkbordercolor={1 1 1}, 
    urlbordercolor={1 1 1},  
    pdfborder={0 0 0}        
}

\usepackage[margin=1in]{geometry} 

\usepackage{amsmath}
\usepackage{bm}
\usepackage{booktabs} 
\usepackage{tabularx} 
\usepackage[titletoc]{appendix}

\definecolor{lightgray}{gray}{0.8}
\definecolor{verylightgray}{gray}{0.95}

\usepackage[utf8]{inputenc} 
\usepackage{booktabs}       
\usepackage{geometry}       

\usepackage{enumitem}
\usepackage{amsmath}
\usepackage{pdflscape} 
\usepackage{ltablex} 

\journal{Computerized Medical Imaging and Graphics}

\begin{document}

\begin{frontmatter}

\title{Class Balancing Diversity Multimodal Ensemble for Alzheimer’s Disease Diagnosis and Early Detection}

\author[label3]{Arianna Francesconi}\ead{arianna.francesconi@unicampus.it}

\author[dott2]{Lazzaro di Biase}\ead{l.dibiase@policlinicocampus.it}

\author[label2]{Donato Cappetta}\ead{d.cappetta@eustema.it}

\author[label2]{Fabio Rebecchi}\ead{f.rebecchi.guest@eustema.it}

\author[label3,label4]{Paolo Soda\texorpdfstring{\corref{cor}}{}}\ead{p.soda@unicampus.it, paolo.soda@umu.se}

\author[label3]{Rosa Sicilia\texorpdfstring{\fnref{contrib}}{}}\ead{r.sicilia@policlinicocampus.it}

\author[label3]{Valerio Guarrasi\texorpdfstring{\fnref{contrib}}{}}\ead{valerio.guarrasi@unicampus.it}

\author[]{for the Alzheimer's Disease Neuroimaging Initiative\texorpdfstring{\fnref{adni}}{}}

\affiliation[label3]{organization={Unit of Computer Systems and Bioinformatics, Department of Engineering, Università Campus Bio-Medico di Roma},
            city={Rome},
            state={Italy}}

\affiliation[dott2]{organization={Operative Research Unit of Neurology, Fondazione Policlinico Universitario Campus Bio-Medico},
            city={Rome},
            state={Italy}}

\affiliation[label2]{organization={Eustema S.p.A., Research and Development Centre},
            city={Naples},
            state={Italy}}

\affiliation[label4]{organization={Department of Diagnostic and Intervention, Radiation Physics, Biomedical Engineering, Umeå University},
            city={Umeå},
            state={Sweden}}

\cortext[cor]{Correspondence: paolo.soda@umu.se, p.soda@unicampus.it}

\fntext[contrib]{These authors contributed equally to this work.}

\fntext[adni]{Data used in preparation of this article were obtained from the Alzheimer's Disease Neuroimaging Initiative (ADNI) database (adni.loni.usc.edu). As such, the investigators within the ADNI contributed to the design and implementation of ADNI and/or provided data but did not participate in analysis or writing of this report. A complete listing of ADNI investigators can be found at: \url{http://adni.loni.usc.edu/wp-content/uploads/how_to_apply/ADNI_Acknowledgement_List.pdf}.}

\begin{abstract}
Alzheimer's disease (AD) poses significant global health challenges due to its increasing prevalence and associated societal costs. Early detection and diagnosis of AD are critical for delaying progression and improving patient outcomes. Traditional diagnostic methods and single-modality data often fall short in identifying early-stage AD and distinguishing it from Mild Cognitive Impairment (MCI). This study addresses these challenges by introducing a novel approach: multImodal enseMble via class BALancing diversity for iMbalancEd Data (IMBALMED). IMBALMED integrates multimodal data from the Alzheimer’s Disease Neuroimaging Initiative database, including clinical assessments, neuroimaging phenotypes, biospecimen and subject characteristics data. It employs an ensemble of model classifiers, each trained with different class balancing techniques, to overcome class imbalance and enhance model accuracy. We evaluate IMBALMED on two diagnostic tasks (binary and ternary classification) and four binary early detection tasks (at 12, 24, 36, and 48 months), comparing its performance with state-of-the-art algorithms and an unbalanced dataset method. IMBALMED demonstrates superior diagnostic accuracy and predictive performance in both binary and ternary classification tasks, significantly improving early detection of MCI at 48-month time point. The method shows improved classification performance and robustness, offering a promising solution for early detection and management of AD.

\end{abstract}


\end{frontmatter}

\section{Introduction}
Alzheimer’s disease (AD) is a neurodegenerative disorder characterized by progressive cognitive deterioration, along with declining activities of daily living and behavioral changes. It primarily affects regions of the brain that are involved in memory, cognition, and language, leading to significant impairment in an individual's ability to engage in conversation and interact with their surroundings~\cite{it1986alzheimer}. AD is the most common form of dementia, affecting over 55 million people worldwide in 2023, with 10 million new cases per year~\cite{who2023} implying a new case diagnosed every three seconds~\cite{world2018global}. The World Health Organization estimates that this number will increase to 78 million people by 2030 and to 139 million by 2050~\cite{shin2022dementia}. In Europe, the social and economic costs of dementia were about €238.6 billion in 2010 and, based on the expected increase in people with dementia, it is estimated to increase to €343 billion by 2050~\cite{maresova2020anticipated}. Considering the escalating global prevalence of AD and the potential economic and societal burdens it entails, there is a pressing need for innovative and effective approaches to address both the diagnosis and early detection of AD.
 
Current available treatments decelerate only the progression of AD and no treatment developed so far is capable of curing a patient with this disease. Thus, developing strategies for detection of AD at early stages, prior to the manifestation of clinical symptoms, is crucial for timely treatment and progression delay~\cite{alhazmi2022update}.
Generally, patients can be categorized into three groups at diagnosis: AD, Mild Cognitive Impairment (MCI), and Cognitively Normal (CN). MCI serves as an intermediate stage between CN and AD, defined to describe individuals experiencing mild cognitive deficits yet still capable of performing daily activities. MCI is further subdivided into Early Mild Cognitive Impairment (EMCI) and Late Mild Cognitive Impairment (LMCI), reflecting different degrees of cognitive decline: EMCI represents milder symptoms, while LMCI involves more severe symptoms similar to early-stage of AD. 
Patients with MCI face a higher risk of progressing to Alzheimer's disease, highlighting the importance of accurately diagnosing MCI and predicting its conversion to AD at an early stage~\cite{education1,iddi2019predicting}. In this context, advancements in artificial intelligence (AI) are increasingly pivotal in enhancing the early detection of MCI~\cite{vrahatis2023revolutionizing,subasi2020use}. These technologies offer the potential to improve diagnostic accuracy and predict disease progression more effectively. However, the application of AI in this field is not without challenges. In particular, class imbalance is one of the most relevant challenges, where the number of patients who convert from MCI to AD is significantly lower than those who do not~\cite{dubey2014analysis}. This imbalance can lead to biased machine learning (ML) models that favor the majority class, typically CN, resulting in a poor identification of potential AD or MCI cases.
Furthermore, to date the AD diagnosis relies primarily on a single data source, also known as modality, such as neuropsychological tests, which assess cognitive abilities like memory, attention, and language that decline with disease progression. However, using a single modality often fails to provide the comprehensive information needed to identify distinguishing patterns for early AD diagnosis, making it challenging for clinicians to detect the disease at its initial stages~\cite{sharma2022comprehensive}. 
In this context, the integration of multiple data sources offers the potential for a more comprehensive and accurate assessment of an individual’s risk of developing AD~\cite{venugopalan2021multimodal}. In addition, biomarkers from various modalities, which are measurable indicators of the severity or presence of some disease state, play a prominent role in the diagnosis of AD. Multiple biomarkers have been explored and investigated through brain structural, neurochemical, and behavioral studies, offering diverse information~\cite{mandal2015brain,mandal2012mapping}. Key biomarkers include brain imaging markers such as Magnetic Resonance Imaging (MRI) and Positron Emission Tomography (PET), Cerebrospinal Fluid (CSF) and genetic markers, such as those in the ApoE gene, which is involved in cholesterol transport and is linked to an increased risk of AD~\cite{namboori2011apoe}. Recognizing the diverse nature of these biomarkers, the integration of multiple modalities offers a promising avenue in the diagnosis and early detection of AD, providing a comprehensive understanding and enhancing AI model accuracy.

To address the challenges related to diverse data modalities and data imbalance, we propose a novel imbalance method: multImodal enseMble via class BALancing diversity for iMbalancEd Data (IMBALMED). IMBALMED employs an ensemble of model classifiers, each trained with different class balancing techniques, and leverages the integration of multimodal tabular data from the Alzheimer’s Disease Neuroimaging Initiative (ADNI)~\cite{ADNI} database. To achieve this goal, we provide the following contributions: 

\begin{itemize}[noitemsep,nolistsep]
    \item Performance Evaluation: we assess IMBALMED's effectiveness in diagnosing AD at the recruitment start, distinguishing between CN and AD (binary classification) as well as CN, AD, and MCI (ternary classification). Additionally, we predict disease progression with binary tasks (AD vs. CN+MCI) at 12, 24, 36, and 48 months, addressing various stages of AD's progression. 
    \item Integration of multimodal data: we harness the complementary information provided by diverse sources, by combining data across four different modalities, i.e., clinical assessments, biospecimen data, subject characteristics, and neuroimaging phenotypes, all obtained from the ADNI database.
    \item Class balancing diversity: we adjust for the diversity in data representation by training an ensemble of models on different class distributions for each modality fostering a comprehensive learning process that captures diverse data characteristics.
    \item Robust comparative analysis in different scenarios: we ensure the robustness and reliability of our methodology, by comparing our approach with an unbalanced dataset approach and nine state-of-the-art algorithms specifically designed to handle imbalanced data. Additionally, we carry out statistical tests to verify that our results are significantly better than our competitors.
\end{itemize}

The rest of this manuscript is organized as follows: the next section provides background on the field of AD, beginning with key challenges and followed by a review of state-of-the-art studies. Section~\ref{sec:Materials} introduces the datasets used in this work and the inclusion criteria for selected ADNI patients, whereas Section~\ref{sec:Methods} presents our method. Section~\ref{sec:Experimental_setup} describes the experimental configuration; Section~\ref{sec:Results} reports and discusses the results. Finally, Section~\ref{sec:Conclusions} provides concluding remarks.

\section{Background}\label{sec:Background}
AI is revolutionizing medical diagnostics, particularly in the detection of AD~\cite{sharma2022comprehensive,venugopalan2021multimodal}. These technologies automate the analysis of complex data, enabling ML models to extract and interpret critical insights, essential for creating effective predictive tools. In this section, we will briefly discuss the primary challenges in the field of AD: the use of multiple data modalities to enhance AI model performance and the class imbalance in datasets, where examples are unevenly distributed across different classes. We will then review some state-of-the-art studies selected based on the following criteria: use of multiple data modalities, methods for addressing the dataset imbalance problem, and data obtained from the ADNI database. This ensures consistency with our selected database and enables meaningful result comparisons. ADNI is the most comprehensive public dataset on Alzheimer's Disease, offering the broadest range of longitudinal tabular data. Additionally, its widespread use in scientific literature facilitates easier comparison of results.

AD's diagnosis with AI can be traced back to the utilization of individual data modalities, commonly known as unimodal approaches. Some of the key unimodal data sources used for AD diagnosis are: MRI, PET, CSF, genetic information, and demographic information~\cite{weiner2017alzheimer}. Whilst unimodal approaches have significantly advanced our knowledge of AD, it is increasingly recognized that a multimodal approach, which integrates multiple modalities, is essential for capturing the full complexity of the disease~\cite{venugopalan2021multimodal}. 
This method, denoted as multimodal fusion, has been shown to surpass the performance of unimodal models in detecting early cognitive impairment and predicting AD conversion~\cite{venugopalan2021multimodal}. Multimodal fusion can occur at various levels, including late, intermediate, and early fusion. Late fusion involves combining the outputs of individual unimodal models, integrating their decisions at a final stage; intermediate fusion combines features at an intermediate layer within the network; early fusion, on the other hand, integrates raw data from different modalities right at the input level. Late fusion offers a significant advantage over others by ensuring a consistent representation of decisions across all modalities and simplifies the combination of multiple modalities at the decision level, making it easier to employ the most optimal analysis methods for each modality independently~\cite{sharma2022comprehensive}. Researchers are continually exploring various fusion methods, seeking the optimal approach to combine diverse data sources for improved classifier performance.

Another ongoing challenge is training AI models with imbalanced datasets~\cite{johnson2019survey}. Models trained on such datasets may achieve high accuracy by predominantly predicting the majority class, but they often struggle to accurately identify the minority class cases. Several techniques are used to address the data imbalance issue; three primary approaches are: undersampling, oversampling, and ensemble learning. Undersampling involves randomly removing data points from classes that have an excess of samples to balance class distributions. Whilst it promotes equal class sizes and faster training times, under-sampling risks discarding potentially informative data. Oversampling, in contrast, generates new synthetic data  data from under-represented classes to equalize class sizes, though it risks introducing overfitting~\cite{brownlee2020random}. Several studies have shown that undersampled datasets yield stable and promising results, compared to oversampled datasets~\cite{dubey2014analysis}. Finally, ensemble learning leverages multiple models to achieve better predictive performance than what could be obtained from any of the models alone. This method has been shown to improve model performance~\cite{fort2019deep}, outperforming single-model approaches in handling imbalanced data~\cite{feng2018class}, despite being less explored~\cite{galar2011overview,aguiar2023survey}.

To the best of our knowledge, there are only a few contributions in the literature that address both class imbalance and multimodal learning on the ADNI dataset to improve Alzheimer's disease diagnosis~\cite{sun2021optimized,velazquez2022multimodal,brand2020task}.
Sun et al.~\cite{sun2021optimized} proposed a new algorithm combining SMOTE~\cite{SMOTE} with a multi-dimensional Gaussian probability density hypothesis~\cite{naseriparsa2020rsmote}, utilizing a Random Forest (RF) classifier, but without explicitly detailing the specific validation method employed. 
Their method integrated data from six modalities, including MRI and PET images, genetics, cognitive tests, CSF, and blood biomarkers at the feature level. 
This approach achieved an accuracy of 95.62\% and a recall of 90.36\% on CN vs. AD classification. Additional results showed an accuracy of 90.82\% and a recall of 80.25\% for EMCI vs. AD, and 85.15\% accuracy with 68.12\% recall for LMCI vs. AD. 
Velazquez et al.~\cite{velazquez2022multimodal} developed a multimodal ensemble model to predict AD conversion in EMCI patients. 
The model leverages two key data sources: Diffusion Tensor Imaging (DTI), which is a type of MRI technique that specifically measures the diffusion of water molecules in brain tissue, and Electronic Health Records (EHR), including patient demographics, genetic biomarkers, and neuropsychological test scores. The model combines a Convolutional Neural Network to process the DTI images with a RF classifier for EHR-based data, employing cross-validation method. A grid search was employed to optimize the weighting between the two classifiers, improving the overall prediction accuracy. Given the inherent class imbalance (fewer patients who convert to AD), oversampling and data augmentation techniques were applied to address the issue, resulting in a 98.81\% accuracy in predicting the conversion from EMCI to AD.
Brand et al.~\cite{brand2020task} proposed the Task Balanced Multimodal Feature Selection (TBMFS) method, which balances feature selection across genetic and neuroimaging data for a binary classification task to distinguish between AD and non-AD (CN and MCI patients).The TBMFS method balances feature selection by jointly optimizing the feature selection process and the classification task through a combined objective function, employing six-fold cross-validation. By integrating matrix factorization and support vector machines (SVM), the TBMFS method achieved a balanced accuracy of 72.8\%, outperforming other models like SVM (59.8\%) and logistic regression (62.9\%). 

As aforementioned, to the best of our knowledge,~\cite{sun2021optimized,velazquez2022multimodal,brand2020task} are the only studies addressing the issue of imbalanced datasets and using multimodal data from the ADNI database. 
This highlights a gap in the AD's field utilizing the ADNI dataset. Most existing balancing methods focus on creating balanced subsets using all samples from the minority class and varying portions of the majority class~\cite{aguiar2023survey}. In contrast, our study aims to explore all possible combinations of minority and majority class samples, under the hypothesis that the diversity of the resulting subsets may provide more informative data for the model and thereby improve performance. Additionally, no literature presented before also evaluate AI models in early detection scenarios, despite the importance of early intervention for improving patient outcomes and slowing disease progression. Furthermore, many studies in the field of AD often rely on DL methods, which, despite their effectiveness, come with a substantial computational cost compared to traditional ML approaches~\cite{schwartz2020green}. In this study, therefore, we focus on utilizing tabular data, including features extracted from images, and integrating it with ML classifiers, as ML methods have been shown to perform better with this type of data~\cite{shwartz2022tabular}.

\section{Materials}\label{sec:Materials}

\subsection{Dataset and Pre-Processing}\label{sec:Dataset}
Data for this study were obtained from the Alzheimer's Disease Neuroimaging Initiative (ADNI) database~\cite{ADNI}, a cornerstone in AD's research, for its broad patient base and diverse data types, crucial for our multimodal analysis. 
ADNI consists of five phases: ADNI 1 (2004-2010), GO (2009-2011), 2 (2011-2016), 3 (2016-2022), and 4 (2022-present). Our study focuses on the first four phases, as ADNI 4 lacks some data modalities essential for our research. Each phase is characterized by specific goals: ADNI 1 aimed to develop biomarkers for clinical trials; ADNI GO investigated biomarkers in early disease stages; ADNI 2 built on this by refining biomarkers as predictors of cognitive decline and outcome measures; and ADNI 3 focused on using functional imaging techniques in clinical trials.

\begin{table}[t]
\begin{adjustbox}{width=\textwidth}
\begin{tabular}{|l|ll|ll|}
\hline
\multicolumn{1}{|c|}{\multirow{2}{*}{\textbf{Modality}}} & \multicolumn{2}{c|}{\textbf{\# of features}} & \multicolumn{2}{c|}{\textbf{\% of missing data}} \\ \cline{2-5}
\multicolumn{1}{|c|}{} & \multicolumn{1}{c|}{\textbf{Diagnostic tasks}} & \multicolumn{1}{c|}{\textbf{Early detection tasks}} & \multicolumn{1}{c|}{\textbf{Diagnostic tasks}} & \multicolumn{1}{c|}{\textbf{Early detection tasks}} \\ \hline
Assessment      & \multicolumn{1}{l|}{37}              & 110 & \multicolumn{1}{l|}{31.06}            & 30.54                                \\ \hline
Biospecimen     & \multicolumn{1}{l|}{1}              & 4 & \multicolumn{1}{l|}{4.48}           & 16.82 \\ \hline
\begin{tabular}[c]{@{}l@{}}Image \\ Analysis\end{tabular} & \multicolumn{1}{l|}{9} & 10 & \multicolumn{1}{l|}{15.34}            & 11.29 \\ \hline
\begin{tabular}[c]{@{}l@{}}Subject \\ Characteristics\end{tabular} & \multicolumn{1}{l|}{6}              & 7 & \multicolumn{1}{l|}{1.22}           & 0.18 \\ \hline
\end{tabular}
\end{adjustbox}
\caption{Summary of selected ADNI data modalities measured at baseline, including the number of features and percentage of missing data for diagnostic and early detection tasks.}
\label{tab:modalities}
\end{table}

\begin{figure}[t]
    \includegraphics[width=\linewidth]{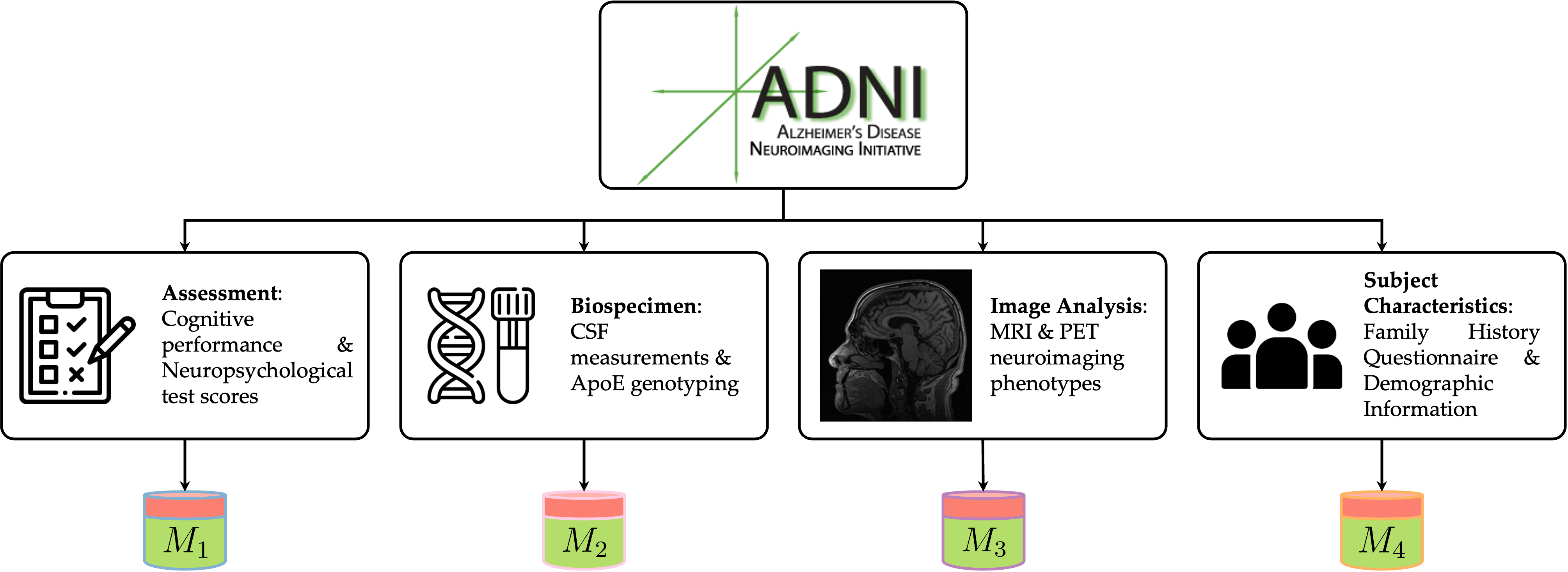}
    \centering
    \caption{Overview of selected data modalities from the ADNI dataset. Each modality is depicted in a labeled box, accompanied by an icon representing the data type and specific tests included in the modality. Arrows indicate the flow of data from the ADNI database to its integration into our final dataset. For notation, see Section~\ref{sec:balancing}.}
    \label{fig:grafico_modalities}
\end{figure}

We utilized four data modalities from the ADNI dataset collected at the start of the recruitment, referred to as the baseline: Assessment, Biospecimen, Image analysis, and Subject Characteristics. A visual representation of these selected modalities is shown in Figure~\ref{fig:grafico_modalities}. Table~\ref{tab:modalities} provides a summary of the number of features and the percentage of missing data for each modality. Furthermore, a detailed list of selected features for each modality and task is provided in Table~\ref{tab:features_selected} in~\ref{app:features}. The Assessment modality includes cognitive performance measures from the Functional Activities Questionnaire, Geriatric Depression Scale, and Modified Hachinski Ischemia Scale, along with neuropsychological test scores such as the Alzheimer’s Disease Assessment Scale, Clinical Dementia Rating, (Modified) Preclinical Alzheimer’s Cognitive Composite, and Mini-Mental State Examination. The Biospecimen modality includes CSF measurements, ApoE genotyping, and laboratory data (blood, urine, and chemistry panels). Genetic data like Genome-Wide Association Studies and Whole Genome Sequencing were excluded due to format inconsistencies with our tabular data needs. Image analysis, derived from MRI and PET scans, provides structural and functional brain data, including key neuroimaging biomarkers like hippocampal volume, entorhinal cortical thickness, and glucose metabolism in specific brain regions. The Subject Characteristics modality includes the Family History Questionnaire (parental and sibling history of AD) and demographic data (education, gender, age, weight, dominant hand, and birth month/year).

The selected modalities were pre-processed as follows: initially, features with more than 50\% missing data were excluded; we then applied One-Hot Encoding for categorical variables and normalized the data to ensure uniform scale across different modalities; missing values were imputed with the k-NN algorithm with k set to 5. Normalization and imputation steps were carefully differentiated between the training set and the validation/test sets. For the training set, we calculated normalization parameters, i.e., mean and standard deviation, and imputed missing values based on its own data. For the validation and test sets, we used the normalization parameters derived from the training set's pre-processing and applied the imputation method fitted on the train set to ensure consistency across all subsets and avoid introducing bias.

\subsection{Patient Selection}\label{sec:ADNI Participants Selection}
Our analysis involved patients from the ADNI study and we chose patients from the ADNI 1, GO, 2, and 3 phases. For the diagnostic tasks, we differentiated between CN, AD, and MCI through binary (CN vs. AD) and ternary (CN vs. AD vs. MCI) classifications, reflecting real-world clinical differentiation challenges. The early detection tasks focused on predicting disease progression by classifying AD vs. CN+MCI, grouping CN and MCI patients into a single class. This classification reflected whether treatment would be required (AD patients) or not (CN+MCI patients) at four future time points (12, 24, 36, and 48 months) from recruitment. The goal was to identify patients at risk of developing AD who could benefit from pharmacological therapy to slow disease progression.
Table~\ref{tab:paz_reclutati} shows the distribution of patients across different tasks and time points, highlighting the patient classification at baseline and subsequent classifications relevant to the early detection tasks. For these tasks, baseline groups included CN, EMCI, and LMCI classes, whereas patients diagnosed with AD at baseline were excluded from the early detection analysis due to their non-conversion status.

\begin{table}[t]
\begin{adjustbox}{width=\textwidth}
\begin{tabular}{|l|ll|llll|}
\hline
\multicolumn{1}{|c|}{\multirow{2}{*}{\textbf{Patient class}}} & \multicolumn{2}{c|}{\textbf{Diagnostic tasks}} & \multicolumn{4}{c|}{\textbf{Early detection tasks}}                \\ \cline{2-7} 
\multicolumn{1}{|c|}{} & \multicolumn{1}{c|}{\textbf{Binary}} & \multicolumn{1}{c|}{\textbf{Ternary}} & \multicolumn{1}{c|}{\textbf{12 months}} & \multicolumn{1}{c|}{\textbf{24 months}} & \multicolumn{1}{c|}{\textbf{36 months}}            & \multicolumn{1}{c|}{\textbf{48 months}} \\ \hline
CN & \multicolumn{1}{l|}{542} & 542 & \multicolumn{1}{l|}{\multirow{2}{*}{1224}} & \multicolumn{1}{l|}{\multirow{2}{*}{963}} & \multicolumn{1}{l|}{\multirow{2}{*}{659}} & \multirow{2}{*}{549}           \\ \cline{1-3}
MCI          & \multicolumn{1}{l|}{-}               & 1113                                   & \multicolumn{1}{l|}{} & \multicolumn{1}{l|}{}                     & \multicolumn{1}{l|}{}                     &                                \\ \hline
AD           & \multicolumn{1}{l|}{411}             & 411                                   & \multicolumn{1}{l|}{106}                   & \multicolumn{1}{l|}{196}                  & \multicolumn{1}{l|}{197}                  & 144                            \\ \hline
Total        & \multicolumn{1}{l|}{953}             & 2066                                   & \multicolumn{1}{l|}{1340}                  & \multicolumn{1}{l|}{1159}                 & \multicolumn{1}{l|}{856}                  & 693                            \\ \hline
\end{tabular}
\end{adjustbox}
\caption{Distribution of patients for the diagnostic and early detection tasks.}
\label{tab:paz_reclutati}
\end{table}

\section{Methods}\label{sec:Methods}
\begin{figure}[t]
    \includegraphics[width=\linewidth]{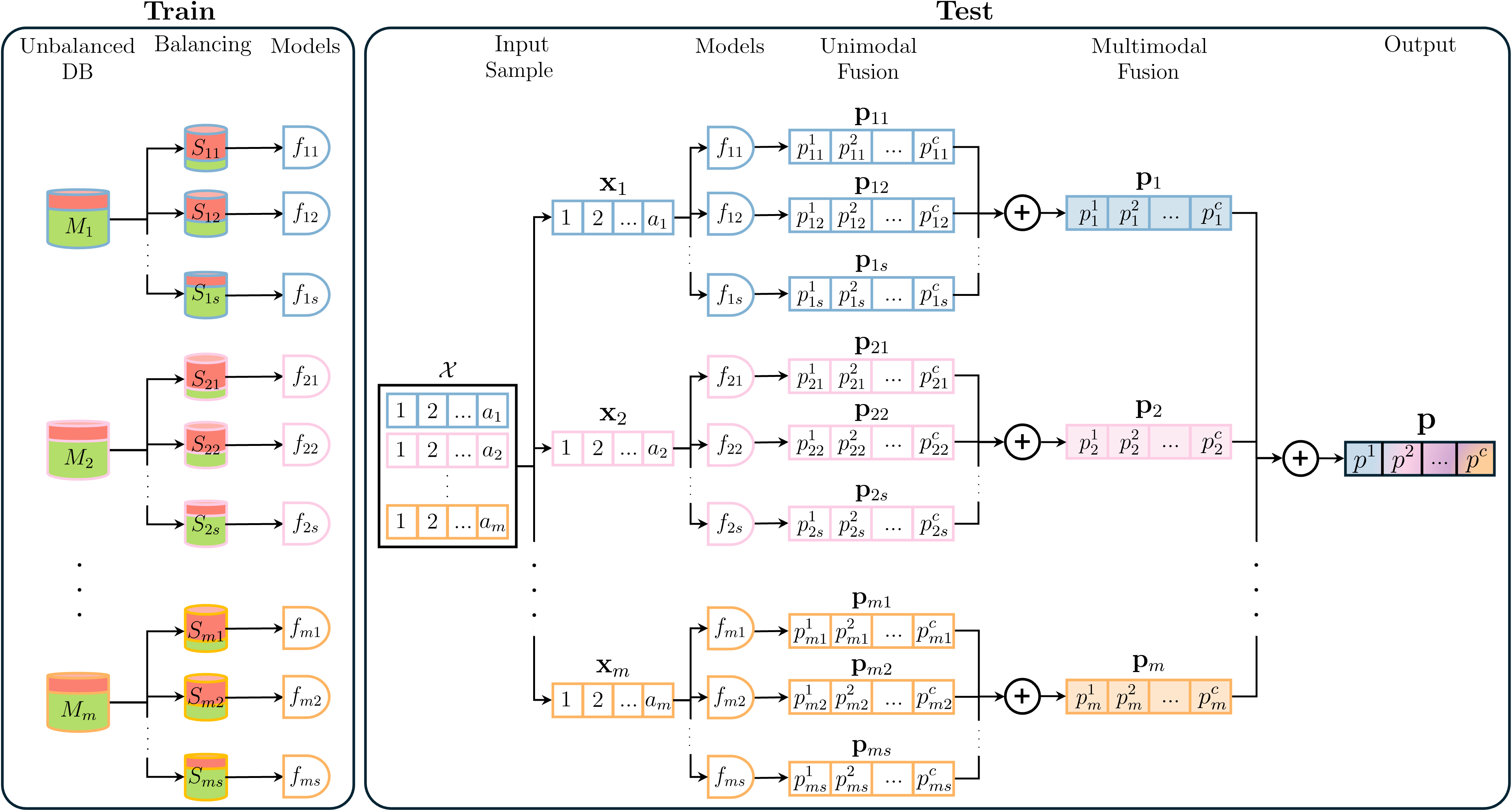}
    \centering
    \caption{Schematic representation of the proposed method, illustrating two main steps: training, which comprises data balancing and classifier training, and testing, which involves unimodal and multimodal fusion to determine class membership probabilities.}
    \label{fig:methods}
\end{figure}

This section details the stages of our method, IMBALMED, also depicted in Figure~\ref{fig:methods} distinguishing between the training and testing stages. For notation, we will use the following conventions: bold lowercase letters for vectors, italics uppercase letters for sets, uppercase letters for matrices, and lowercase letters for scalars.
In the training phase (Train block in Figure~\ref{fig:methods}), data balancing involves creating distinct subsets for each modality, differentiated by their class representativeness. This process creates diverse model experts, where the diversity is driven by varying levels of class representativeness. Next, each balanced subset undergoes training using a classifier. During the testing phase (Test block in Figure~\ref{fig:methods}), given an input sample \( \mathcal{X} \), a set of vectors each characterized by different lengths based on the number of features in each modality, class membership probabilities \( \mathbf{p}_{ij} \) are calculated, where \( i \) refers to the modality and \( j \) to the balanced subset. The unimodal fusion then combines these probability distributions within each modality, producing unimodal probability vectors \( \mathbf{p}_{i} \). Finally, the multimodal fusion averages the unimodal probabilities across modalities, yielding the output vector \( \mathbf{p} \), which represents the probability of class membership for the sample based on the knowledge learned from the various modalities.

The presented methodology introduces a novel multimodal ensemble-based approach that manipulates class distributions during the training of classifiers for each modality. This methodological innovation not only aims to enhance the robustness and performance of individual classifiers but also focuses on their effective integration. 

\subsection{Data Balancing}\label{sec:balancing}
In this study, we address a multimodal classification task with \( m \) modalities using an unbalanced dataset. Given a finite set of modalities \( \mathcal{M} = \{M_1, M_2, \ldots, M_m\} \), where each \( M_i \) represents a distinct data modality for \( i = 1, \ldots, m \), let \( \mathcal{Y} = \{y_1, y_2, \ldots, y_c\} \) represent the set of \( c \) classes with \( y_k \) as the \( k \)-th class for \( k = 1, \ldots, c \). To mitigate the challenge of data imbalance and promote classifier diversity, we employed an undersampling strategy. This involved creating the set \( {S}_i \) of the balanced subsets of the modality \( M_i \), defined as \( \mathcal{S}_i = \{S_{i1}, S_{i2}, \ldots, S_{is}\} \), where \( i \) is the modality index and \( s \) is the number of the balanced subset within that modality. By creating the subsets, we aim to introduce diversity to the trained model through varying levels of class balance. For each balanced subset \( S_{ij} \), with \( j = 1, \ldots, s \), the combination of class representativeness levels is defined by the vector:
\begin{equation}
    \mathbf{b_{ij}} = [b_{ij}^1, b_{ij}^2, \ldots, b_{ij}^c], \quad \mathbf{b_{ij}} \in \mathbb{R}^c, \quad b_{ij}^k \in \mathbb{R}
\end{equation}
where each element of the vector \( \mathbf{b_{ij}} \) represents the percentage of representativeness of the \( k \)-th class for the subset \( S_{ij} \), scaled between 0 and 1. To ensure significant representation for each class, we define \( r \) as the minimum representativeness number of samples for each class, expressed as a percentage between 0 and 1.  This value \( r \) is the increment by which the representativeness percentages vary from the specified \( r \) to 1, which corresponds to the case where only one class is present in the dataset. Formally, we have: \( 0 < r < 1 \quad \forall \, r \in \mathbb{R}.\)

Each representativeness level \( b_{ij}^k \) must satisfy the following conditions:
\begin{enumerate}[label=\roman*.]

    \item Representativeness levels: Each representativeness \( b_{ij}^k \) is defined by:
    \begin{equation}
    b_{ij}^k = w_{ij}^k \cdot r, \quad w_{ij}^k \in \mathbb{N}, r \in \mathbb{R}
    \end{equation}
    where \( w_{ij}^k \) is a coefficient that scales the minimum representativeness level \( r \) for class \( k \). The values \( w_{ij}^k \cdot r \) are selected from the set \( \mathit{R_c} \), which represents the set of all possible representativeness levels for all classes. For each balanced subset \( S_{ij} \), we define:
    \begin{equation}
    \mathbf{b_{ij}} = [w_{ij}^1, w_{ij}^2, \ldots, w_{ij}^c] \cdot r \in \mathit{R_c} = \left\{ r, 2r, \ldots, \left\lfloor \frac{1}{r} - c + 1 \right\rfloor \cdot r \right\}
    \end{equation}
    
    where \( \left\lfloor \frac{1}{r} - c + 1 \right\rfloor \) represents the maximum representativeness of the specific class within the balanced subset \( S_{ij} \) for multi-class problems with an odd number \( c \) of classes. This value is obtained by calculating the complement to 1 after assigning representativeness values to the other \( c -1 \) classes.

    \item Sum of Representativeness: For each class \( k \), \( b_{ij}^k \) represents the percentage of representativeness for that class. The sum of the representativeness percentages across all classes within the same balanced subset \( \mathbf{b_{ij}} \) must equal 1:
    \begin{equation}
    \sum_{k=1}^c b_{ij}^k = \sum_{k=1}^c w_{ij}^k \cdot r = 1
    \end{equation}

\end{enumerate}

A binary problem ($\mathit{c=2}$) would have $\mathit{s=9}$ balanced subsets ($\mathit{c^3+1}$), whilst a ternary problem ($\mathit{c=3}$) would consist of $\mathit{s=28}$ different balanced subsets. 
To better clarify the methodology used, in the case of a binary problem with $\mathit{r}$ set at 0.1, the computed representativeness levels were: \( \mathbf{b_{i1}}=[0.1, 0.9],  \mathbf{b_{i2}}=[0.2, 0.8], \ldots, \mathbf{b_{i9}}=[0.9, 0.1] \). 
For a ternary problem with $\mathit{r}=0.11 $, the computed representativeness levels were: \( \mathbf{b_{i1}}=[0.11, 0.11, 0.78],  \mathbf{b_{i2}}=[0.11, 0.22, 0.67], \ldots, \mathbf{b_{i28}}=[0.77, 0.11, 0.12] \). 
This strategy ensured that each class maintained a minimum representativeness percentage of $\mathit{r}$ and a number of samples for each class equal to the representativeness level of the class in question for the minimum number of samples among the classes.

\subsection{Data Fusion Levels}\label{sec:Data Fusion Levels}
Our framework incorporates a two-level fusion process, first at the unimodal level and then across different modalities, thereby leveraging the strengths of diverse data representations.
For each modality \( M_i \in \mathcal{M} \) for \( i = 1, \ldots, m \), we train a set of classifiers \( \mathcal{F}_i = \{f_{i1}, f_{i2}, \ldots, f_{is}\} \), where \( j = 1, \ldots, s \) and \( f_{ij} \) represents the model trained on the balanced subset \( S_{ij} \) of the \( i \)-th modality. Each classifier \( f_{ij} \) maps from the feature space to the class probability space: \( f_{ij} : \mathbb{R}^{a_i} \rightarrow \mathbb{R}^c \), where \( a_i \) denotes the number of features for the \( i \)-th modality. For a single input sample \(\mathcal{X} = \{\mathbf{x}_1, \mathbf{x}_2, \ldots, \mathbf{x}_m\} \), the \(\mathbf{x}_i \in \mathbb{R}^{a_i}\) represents the feature vector for the \( i \)-th modality in input to the model. The probability that the input sample belongs to a specific class is expressed by the vector \( \mathbf{p}_{ij} = f_{ij}(\mathbf{x}_i) \) where \(\mathbf{p}_{ij} \in \mathbb{R}^c\) for \( i = 1, \ldots, m \) and \(  j = 1, \ldots, s\). The vector \(\mathbf{p}_{ij}\) is composed of a probability value for each class and expresses the probability that the input sample belongs to each class for the balanced subset \( S_{ij} \), given the classifier \( f_{ij} \).

For the first fusion level, we computed the average of the summed probability distributions \( \mathbf{p}_{ij} \) produced by the classifiers:
\begin{equation}
 \mathbf{p}_{i} = \frac{1}{s} \sum_{j=1}^s \mathbf{p}_{ij}
\end{equation}
where each element of \( \mathbf{p}_{ij} \) is summed with corresponding elements across all \( j \). The resulting vector \(\mathbf{p}_{i}\) is composed of a probability value for each class and expresses the probability that the input sample belongs to each class for the \( i \)-th modality.

At the second fusion level, we aggregate the \(\mathbf{p}_{i}\) probability distributions from each modality:
\begin{equation}
 \mathbf{p} = \frac{1}{m} \sum_{i=1}^m \mathbf{p}_{i}
\end{equation}
where each element of \(\mathbf{p}_{i}\) is summed with the corresponding elements across all \( i \). The resulting vector \(\mathbf{p}\) contains a probability value for each class, representing the likelihood that the input sample belongs to the specific class, considering all \( m \) modalities.

To obtain the final class prediction for the input \(\mathcal{X}\), we apply the argmax function to the aggregated probability distribution \(\mathbf{p}\), selecting the class with the highest probability:
\begin{equation}
    \hat{c}(\mathcal{X}) = \arg\max_{k \in \mathcal{C}} \mathbf{p}
\end{equation}
The predicted class \( \hat{c}(\mathcal{X}) \) is then used during evaluation to compare with the true class label, assessing the model's performance.

\section{Experimental setup}\label{sec:Experimental_setup}
Our study encompasses both diagnostic and early detection tasks. For the diagnostic tasks, we use \( m=4\) data modalities (Section~\ref{sec:balancing}), excluding neuropsychological test scores to avoid bias, as AD diagnosis often relies on such tests~\cite{weller2018current}. For the early detection tasks, baseline neuropsychological test scores are included to assess treatment necessity.

To address data imbalance, we apply the IMBALMED methodology described in Section~\ref{sec:balancing}, setting the minimum representativeness number of samples to \( r=0.1 \) for the binary tasks and \( r=0.11 \) for the ternary task. These parameters were selected to ensure maximum diversity among the subsets, guaranteeing equal class representation, with \( \mathbf{b_{i5}}=[0.5, 0.5] \) for the binary tasks and \( \mathbf{b_{i16}}=[0.33, 0.33, 0.34] \) for the ternary task (see Section~\ref{sec:balancing} for the notation).

\subsection{Training and Evaluation} \label{subsec:training}
Our training methodology involved a 10-fold cross validation with \( k=10 \) and we evaluated our methodology on 25 different classifiers selected from eight families of learning paradigms. This approach was chosen to demonstrate the robustness and performance of our method across various classification tasks, regardless of the model used. The classifiers were used with default hyperparameters, since their tuning is out of the scope of this work. However, the No Free Lunch Theorem states that, in many cases, tuned parameters do not significantly outperform the default ones~\cite{arcuri2013parameter}. The choice of diverse families of learning paradigms aimed to assess the performance across a wide spectrum of classification techniques. Furthermore, testing our method across various learning paradigm families demonstrates its applicability to experts across different domains, not limited to shallow learning models. Table~\ref{tab:classificatori} lists the names of the classifiers used in this study, divided by families.

\begin{table}[t]
  \centering
  \begin{tabular}{
    |>{\centering\arraybackslash}p{3.5cm} 
    |>{\centering\arraybackslash}p{12cm}| 
  }
    \hline
    \textbf{Family} & \textbf{Classifiers} \\
    \hline
    \multicolumn{1}{|>{\raggedright\arraybackslash}p{3.5cm}|}{Artificial Neural Networks} & \multicolumn{1}{>{\raggedright\arraybackslash}p{12cm}|}{Multilayer Perceptron (MLP).} \\
    \hline
    \multicolumn{1}{|>{\raggedright\arraybackslash}p{3.5cm}|}{Decision Trees} & \multicolumn{1}{>{\raggedright\arraybackslash}p{12cm}|}{Decision Tree (DT), Extra-Tree (ET), and Extra-Trees (ETs).} \\
    \hline
    \multicolumn{1}{|>{\raggedright\arraybackslash}p{3.5cm}|}{Ensemble Methods} & \multicolumn{1}{>{\raggedright\arraybackslash}p{12cm}|}{AdaBoost (AB), Gradient Boosting (GB), Hist Gradient Boosting (HGB), Random Forest (RF), XGBoost (XGB), and XGBoost Random Forest (XGBRF).} \\
    \hline
    \multicolumn{1}{|>{\raggedright\arraybackslash}p{3.5cm}|}{Label Propagation} & \multicolumn{1}{>{\raggedright\arraybackslash}p{12cm}|}{Label Propagation (LP).} \\
    \hline
    \multicolumn{1}{|>{\raggedright\arraybackslash}p{3.5cm}|}{Label Spreading} & \multicolumn{1}{>{\raggedright\arraybackslash}p{12cm}|}{Label Spreading (LS).} \\
    \hline
    \multicolumn{1}{|>{\raggedright\arraybackslash}p{3.5cm}|}{Linear Models} & \multicolumn{1}{>{\raggedright\arraybackslash}p{12cm}|}{Linear Discriminant Analysis (LDA), Linear Support Vector (LSVC), Logistic Regression (LR), Passive-Aggressive (PA), Perceptron (Percep.), Ridge (Ridge), Stochastic Gradient Descent (SGD), Support Vector Classification (SVC) and Nu-Support Vector Classification (NuSVC).} \\
    \hline
    \multicolumn{1}{|>{\raggedright\arraybackslash}p{3.5cm}|}{Naive Bayes} & \multicolumn{1}{>{\raggedright\arraybackslash}p{12cm}|}{Bernoulli Naive Bayes (BNB) and Gaussian Naive Bayes (GNB).} \\
    \hline
    \multicolumn{1}{|>{\raggedright\arraybackslash}p{3.5cm}|}{Nearest Neighbors} & \multicolumn{1}{>{\raggedright\arraybackslash}p{12cm}|}{k-Nearest Neighbors (k-NN) and Nearest Centroid (NC).} \\
    \hline
  \end{tabular}
  \caption{Selected classifiers divided by families.}
  \label{tab:classificatori}
\end{table}

For evaluation, we used the G-mean, which ensures both classes are adequately represented and prevents the model from favoring the majority class, making it highly suitable for evaluating models on imbalanced datasets.
For binary tasks, G-mean was calculated as the square root of the product of recall and specificity, whilst for ternary tasks, it was calculated as the cube root of the product of the recall and specificity for each class.

\subsection{Competitors and Statistical Analysis} \label{subsec:competitor}

\begin{table}[t]
  \centering
  \small
  \begin{tabular}{|>{\centering\arraybackslash}p{7cm}|>{\centering\arraybackslash}p{8.5cm}|}
    \hline
    \textbf{Algorithm} & \textbf{Description} \\
    \hline
    \multicolumn{1}{|p{7cm}|}{Adaptive Random Forest (ARF)~\cite{ARF}} & \multicolumn{1}{p{8.5cm}|}{It utilizes the Hoeffding Tree as its base classifier and employs mechanisms to adapt to changes in data distribution over time, known as concept drift.} \\
    \hline
    \multicolumn{1}{|p{7cm}|}{\raggedright Adaptative Random Forest with Resampling (ARFR)~\cite{ARFR}} & 
    \multicolumn{1}{p{8.5cm}|}{\raggedright It is an extension of the ARF that incorporates resampling techniques to handle imbalanced data streams.} \\
    \hline
    \multicolumn{1}{|p{7cm}|}{Comprehensive active learning method for multiclass imbalanced streaming data with concept drift (CALMID)~\cite{CALMID}} & \multicolumn{1}{p{8.5cm}|}{It combines active learning with strategies to address multiclass imbalance and concept drift in data streams, focusing on efficiently selecting the most informative samples for learning.} \\
    \hline
    \multicolumn{1}{|p{7cm}|}{Leveraging Bagging Adwin (LB)~\cite{LB}} & \multicolumn{1}{p{8.5cm}|}{It leverages the Bagging algorithm with the ADaptive WINdowing (ADWIN~\cite{ADWIN}) change detector to handle data with concept drift.} \\
    \hline
    \multicolumn{1}{|p{7cm}|}{Online AdaBoost (OADA)~\cite{OADA}} & \multicolumn{1}{p{8.5cm}|}{It is an online version of the AdaBoost algorithm; it updates the model with new data, adjusting weights of incorrectly classified instances to improve performance.} \\
    \hline
    \multicolumn{1}{|p{7cm}|}{Online Bagging with Adwin (OBA)~\cite{OBA}} & \multicolumn{1}{p{8.5cm}|}{It combines the OB algorithm with the ADWIN algorithm for detecting and adapting to concept drift.} \\
    \hline
    \multicolumn{1}{|p{7cm}|}{Oversampling Online Bagging (OOB)~\cite{OOB}} & \multicolumn{1}{p{8.5cm}|}{It is an adaptation of the Online Bagging algorithm (OB~\cite{OB}) that incorporates oversampling techniques, by focusing on increasing the representation of minority classes.} \\
    \hline
    \multicolumn{1}{|p{7cm}|}{\raggedright Robust Online Self-Adjusting Ensemble (ROSE)~\cite{ROSE}} & \multicolumn{1}{p{8.5cm}|}{\raggedright It dynamically adjusts its composition and parameters in response to concept drift.} \\
    \hline
    \multicolumn{1}{|p{7cm}|}{Undersampling Online Bagging (UOB)~\cite{UOB}} & \multicolumn{1}{p{8.5cm}|}{It extends OB algorithm by incorporating undersampling techniques, reducing the influence of majority class samples to balance the training process.} \\
    \hline
  \end{tabular}
  \caption{Selected competitors.}
  \label{tab:competitors}
\end{table}

We compared our balancing method, IMBALMED, with two main approaches: using the raw, unbalanced dataset without any balancing (unbalanced dataset approach), and employing state-of-the-art data balancing algorithms. From a comprehensive review assessing cutting-edge algorithms~\cite{aguiar2023survey}, we selected the top-performing ensemble-based algorithms for multiclass scenarios, listed in Table~\ref{tab:competitors}. To ensure a rigorous comparison, we evaluated the performance on the test set of these algorithms against our most effective classifier, which was selected based on performance on an independent validation set. Each algorithm was implemented with its default classifier and parameters, in order to adopt the same setup described in~\ref{subsec:training}. 
Furthermore, we assessed the statistical significance of our method IMBALMED against both the unbalanced dataset approach and state-of-art balancing data competitors using the Paired T-test ($p-value \leq 0.05$). Given the cross-validation performance values on the test set, we calculated the win rate, representing the instances where IMBALMED's $p-value$ outperformed the specific competitor. To provide a comprehensive view of IMBALMED's effectiveness and consistency, we also measured the tie rate and loss rate, indicating the percentage of times that our method was equal to or worse than the competitors, respectively.

As mentioned in section~\ref{sec:Background}, there are three related studies addressing imbalance and multimodal learning on the ADNI dataset~\cite{sun2021optimized, velazquez2022multimodal, brand2020task}. Although they could be valid benchmark to assess our approach, these works were excluded from our comparison due to insufficient details for a fair experiments reproduction.
It is also worth noting that we could not validate the proposed approach on external datasets since the literature does not provide publicly available datasets that comply with all modalities selected in this study.

\section{Results and Discussions}\label{sec:Results}
This section presents the evaluation of our proposed multimodal ensemble-based approach, IMBALMED, across the various diagnostic and early detection tasks. We compare the performance of our method against both an unbalanced dataset approach and state-of-the-art algorithms designed for handling imbalanced data. The following subsections provide detailed analyses of the performance outcomes, including comparative statistical evaluations.

\paragraph{Multimodal Fusion}
Table~\ref{tab:our_results} summarizes the performance of our approach across all downstream tasks. It highlights the \textquotedblleft Best G-mean\textquotedblright{} achieved by the top-performing classifier on test set (specified in the column \textquotedblleft Best Classifier\textquotedblright{}) in each experiment, with the associated standard deviation indicating variability across the cross-validation folds. Additionally, the table presents the \textquotedblleft Average G-mean\textquotedblright{}, calculated as the mean G-mean performance across the cross-validation folds and averaged over all classifiers, along with the corresponding standard deviation, reflecting variability in the overall performance of the selected classifiers.

\begin{table}[t]
    \centering
    \begin{tabular}{|l|l|l|l|}
    \hline
    \multicolumn{1}{|c|}{\textbf{Task}} & \multicolumn{1}{c|}{\textbf{Best G-mean}} & \multicolumn{1}{c|}{\textbf{Best Classifier}} & \multicolumn{1}{c|}{\textbf{Average G-mean}} \\ \hline
    Binary diagnostic & 97.40 ± 1.24 & RF & 96.01 ± 1.33 \\ \hline
    Ternary diagnostic & 73.49 ± 4.31 & LR & 69.56 ± 4.38 \\ \hline
    12-month early detection & 79.60 ± 3.45 & AB & 76.10 ± 1.22 \\ \hline
    24-month early detection & 86.10 ± 3.60 & BNB & 83.01 ± 1.03 \\ \hline
    36-month early detection & 83.30 ± 3.47 & AB & 81.24 ± 0.88 \\ \hline
    48-month early detection & 87.24 ± 2.76 & BNB & 85.17 ± 0.86 \\ \hline
    \end{tabular}
    \caption{Data fusion performance: top-performing classifier (\textquotedblleft Best G-mean\textquotedblright{} of the \textquotedblleft Best Classifier\textquotedblright{}) and average performance across classifiers (\textquotedblleft Average G-mean\textquotedblright{}) based on G-mean for each task.}
    \label{tab:our_results}
\end{table}

In the diagnostic task, the binary task achieves the greatest robustness in performance, with a low standard deviation of 1.24 in the average G-mean value, whilst the ternary task exhibits greater performance variation, reflecting the increased difficulty of this classification task.
In the early detection tasks, performance varied slightly across different time horizons. Both the highest performance and the lowest standard deviation are achieved in the 48-month task. This is because, at this time point, most of the MCI patients have converted to AD, making the task easier. These results highlight the robustness and adaptability of the proposed framework in handling different types of tasks and time horizons. The low standard deviations in average G-mean values across all tasks, especially in the 36-month and 48-month early detection tasks, further emphasize the stability and reliability of the approach.
Notably, three out of the six tasks are best performed by ensemble methods, highlighting the power of leveraging a combination of multiple classifiers also known as base learners. This finding highlights their ability to enhance generalizability in complex scenarios and the power of our multimodal ensemble-based approach, which utilizes ensemble techniques as a key component in our framework.

From the unimodal fusion results, as reported in Table~\ref{tab:unimodal_results} of the \ref{app:unimodal}, it is evident that the Assessment modality performs most effectively across all tasks. In diagnostic tasks, the performance difference between the Assessment modality and the second most effective modality (Image Analysis) is approximately 11.6\% for the binary task and 34.8\% for the ternary task. In early detection tasks, the performance difference between the two modalities is 4.98\%, 2.31\%, 3.37\%, and 1.97\% for the 12-, 24-, 36-, and 48-month tasks, respectively. The smaller performance difference between Assessment and other modalities in early detection tasks is due to the increased complexity of the task, benefiting more from multimodal data. The smallest difference is observed in the clinically most complex 48-month task: this extended prediction period introduces greater uncertainty and variability, making the task more challenging compared to the 12-, 24-, and 36-month tasks.

\paragraph{Competitors and Statistical Analysis}
\begin{figure}[ht]
    \centering
    \begin{subfigure}[b]{0.48\textwidth}
        \centering
        \includegraphics[width=\textwidth]{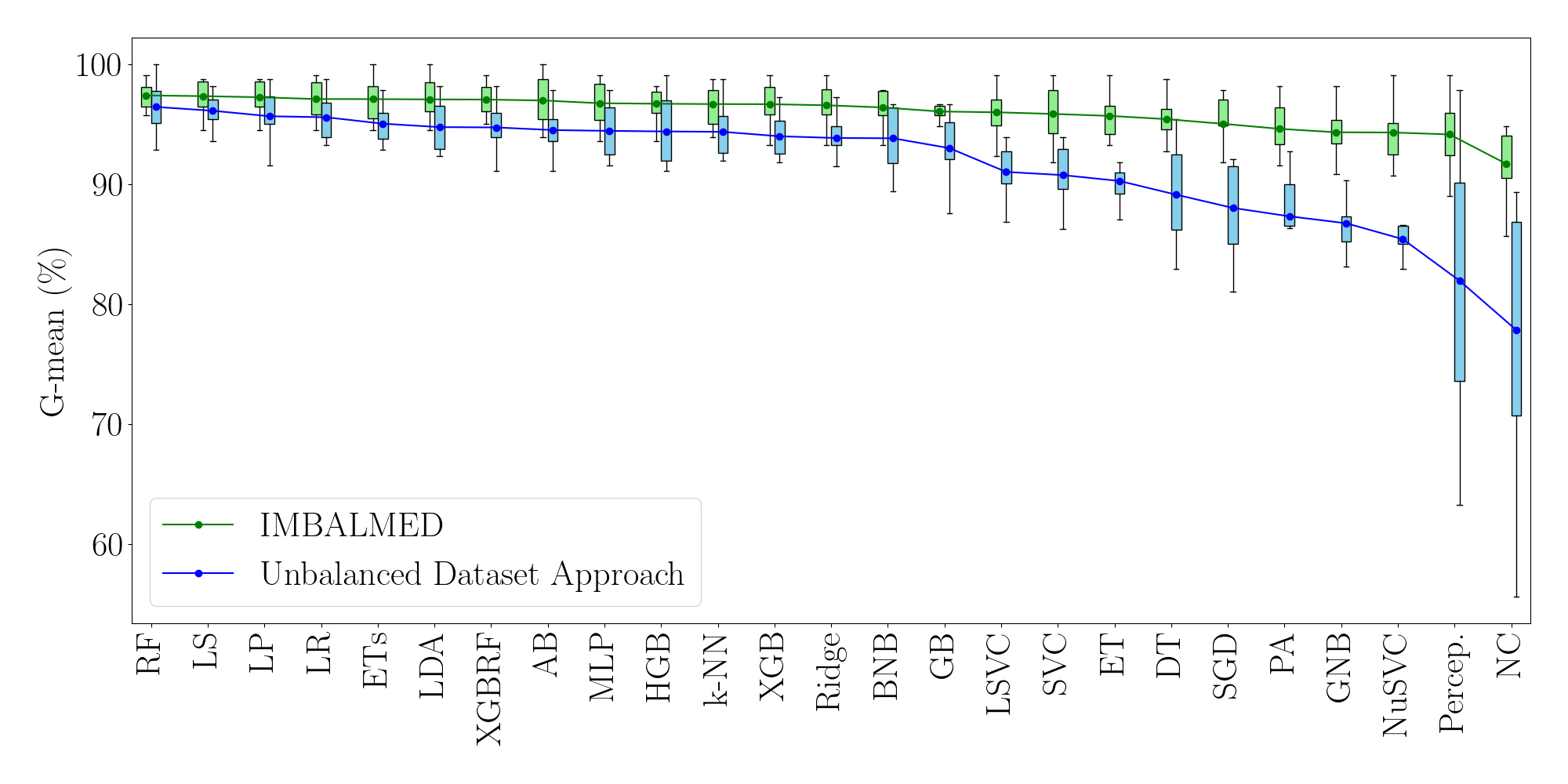}
        \caption{Binary diagnostic task.}
        \label{fig:binary}
    \end{subfigure}
    \begin{subfigure}[b]{0.48\textwidth}
        \centering
        \includegraphics[width=\textwidth]{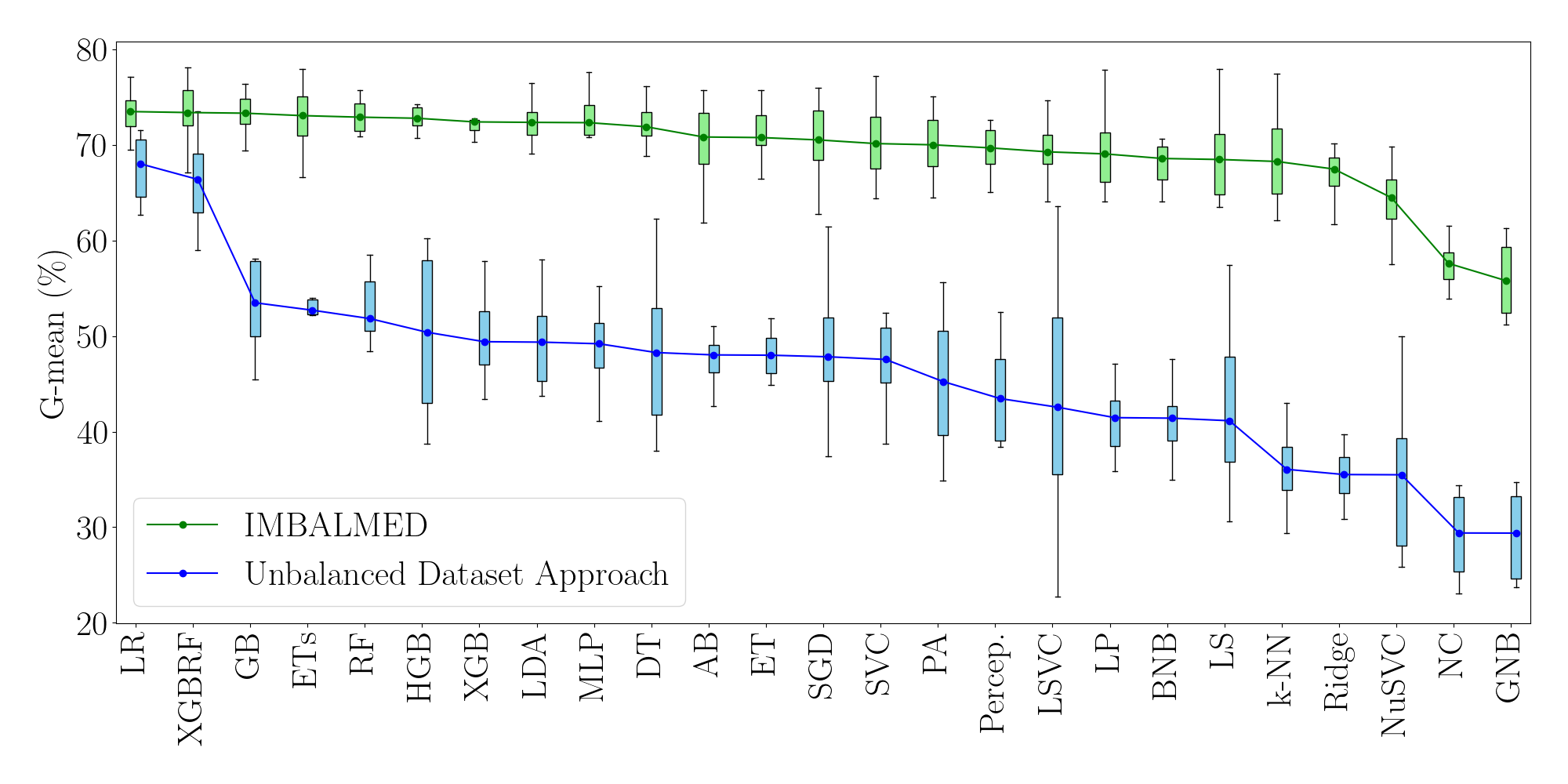}
        \caption{Ternary diagnostic task.}
        \label{fig:ternary}
    \end{subfigure}

    \vspace{3mm} 

    \begin{subfigure}[b]{0.48\textwidth}
        \centering
        \includegraphics[width=\textwidth]{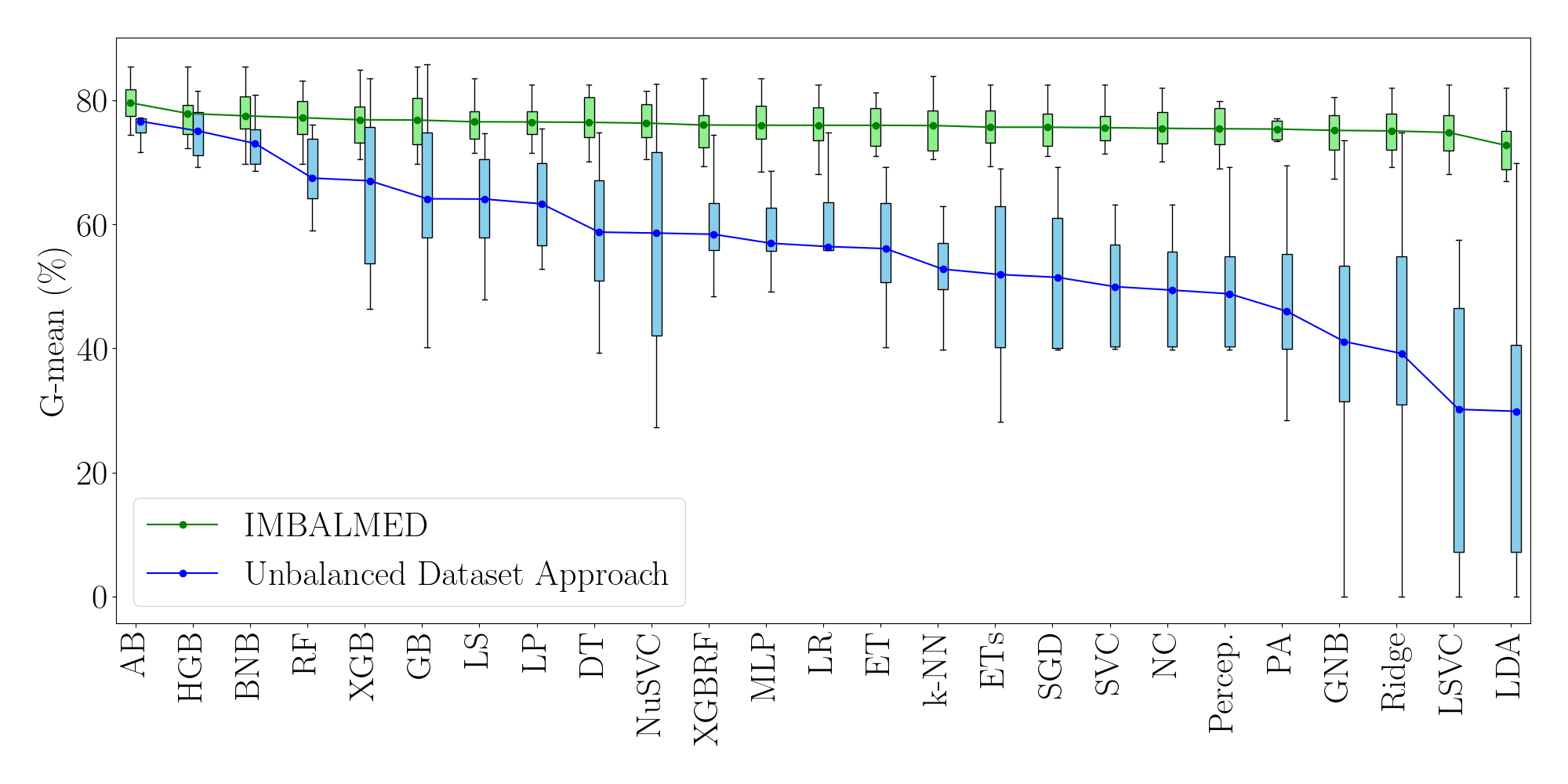}
        \caption{12-month early detection task.}
        \label{fig:12progn}
    \end{subfigure}
    \hspace{1mm}
    \begin{subfigure}[b]{0.48\textwidth}
        \centering
        \includegraphics[width=\textwidth]{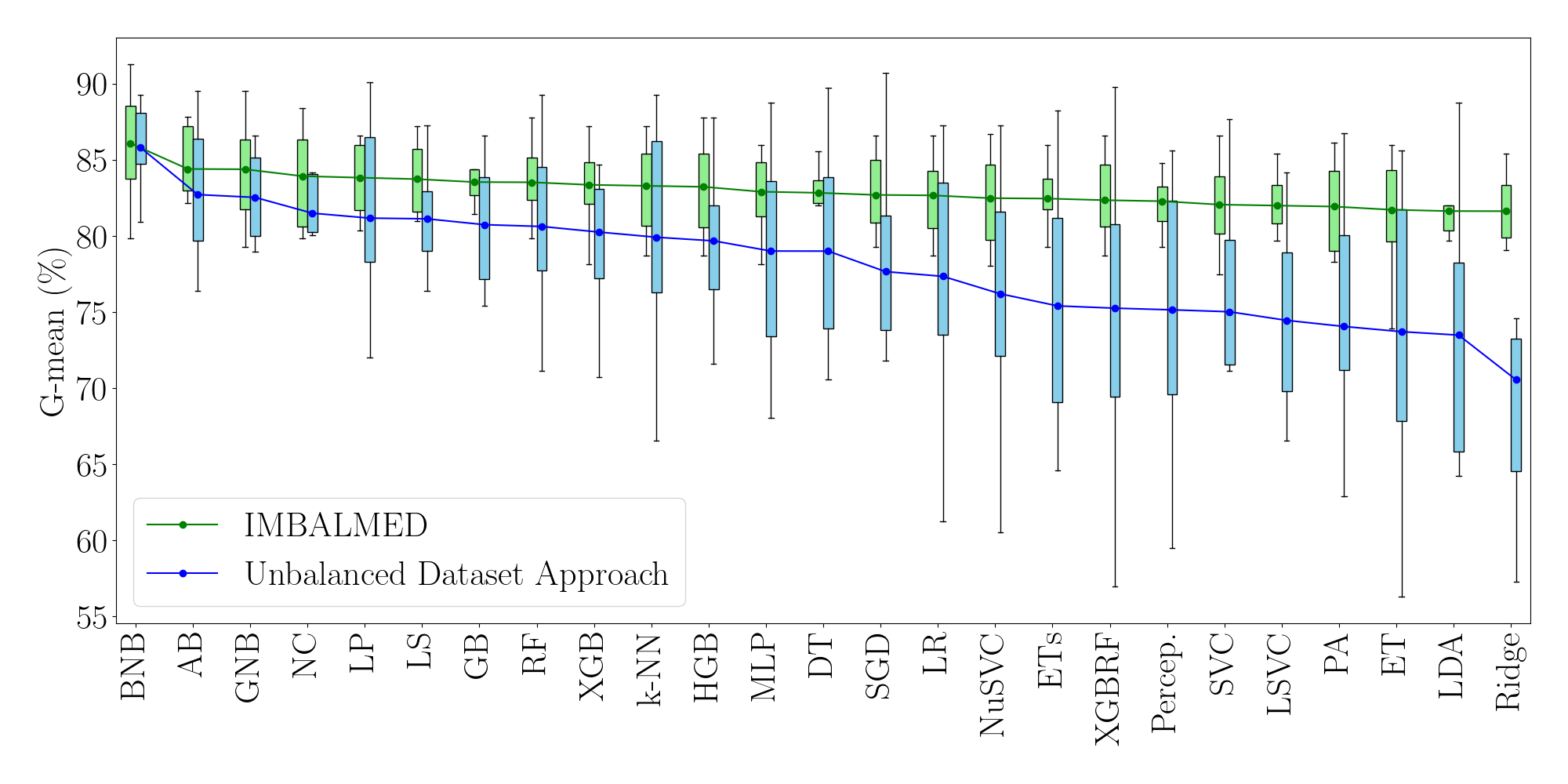}
        \caption{24-month early detection task.}
        \label{fig:24progn}
    \end{subfigure}

    \vspace{3mm} 

    \begin{subfigure}[b]{0.48\textwidth}
        \centering
        \includegraphics[width=\textwidth]{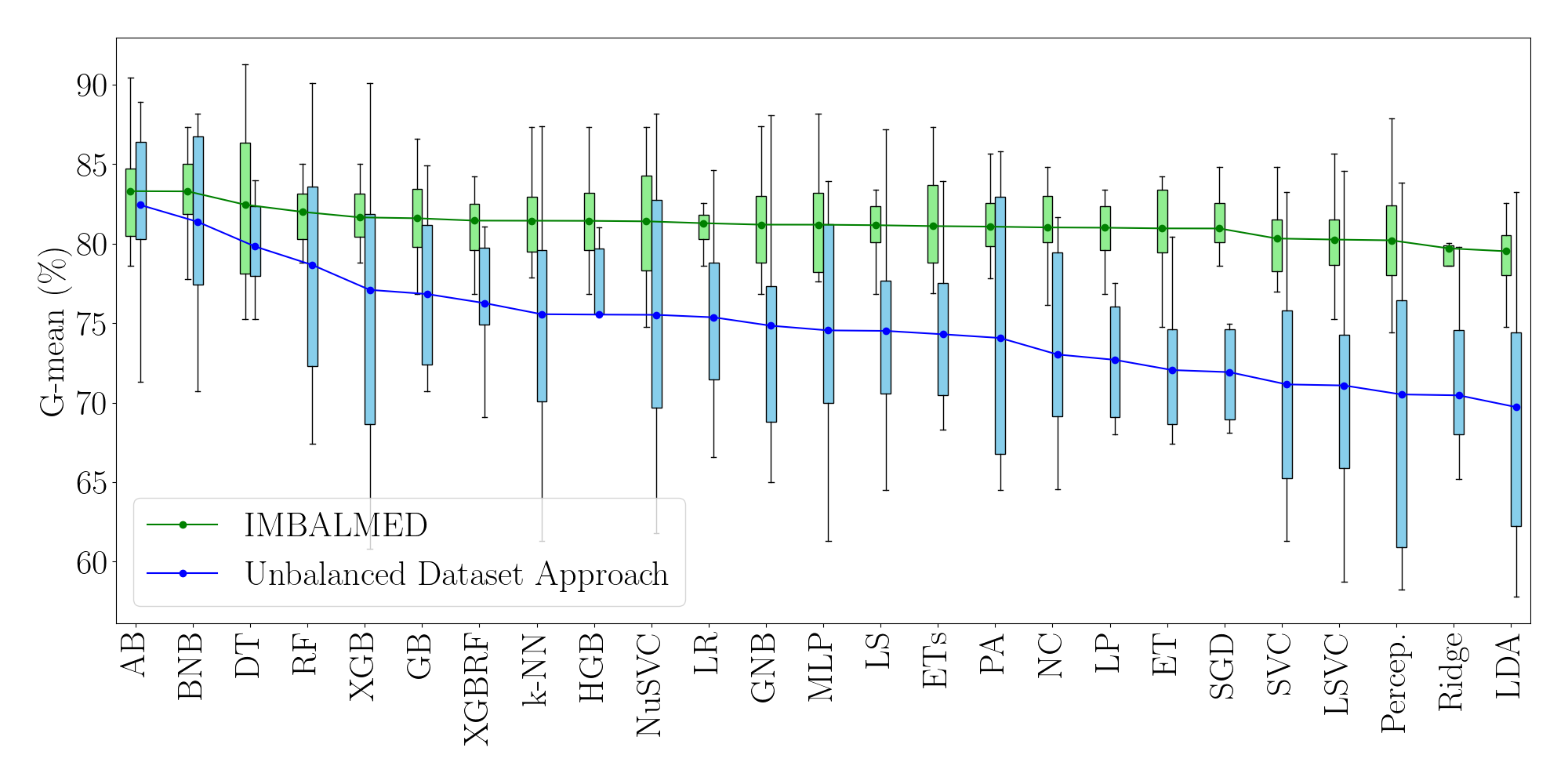}
        \caption{36-month early detection task.}
        \label{fig:36progn}
    \end{subfigure}
    \hspace{1mm}
    \begin{subfigure}[b]{0.48\textwidth}
        \centering
        \includegraphics[width=\textwidth]{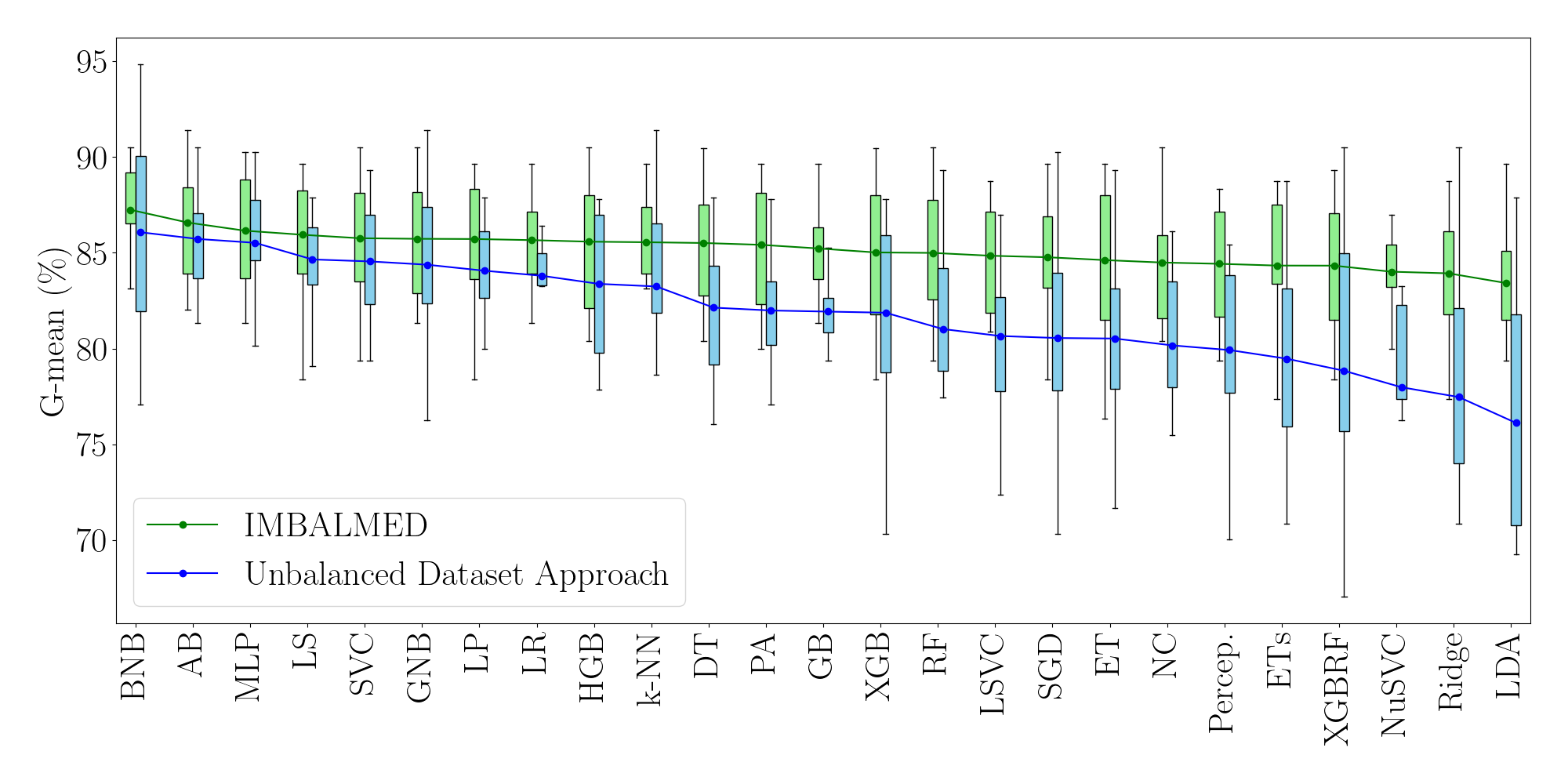}
        \caption{48-month early detection task.}
        \label{fig:48progn}
    \end{subfigure}

    \caption{IMBALMED vs. Unbalanced Dataset Approach for various tasks. Panels (a) and (b) represent the binary and ternary diagnostic tasks, respectively. Panels (c) to (f) represent the 12-, 24-, 36-, and 48-month early detection tasks, respectively. Lines connect classifiers' average performance.}

    \label{fig:combined_unbalancedDB}
    
\end{figure}

The performance obtained with our methodology was first compared with the performance achieved using the unbalanced dataset approach. Figure~\ref{fig:combined_unbalancedDB} displays the results of these comparisons for the diagnostic tasks (Figures~\ref{fig:binary} and~\ref{fig:ternary}) and the early detection tasks at 12-month (Figure~\ref{fig:12progn}), 24-month (Figure~\ref{fig:24progn}), 36-month (Figure~\ref{fig:36progn}) and 48-month (Figure~\ref{fig:48progn}). 
Each figure contains 25 boxplots per method, corresponding to the 25 selected classifiers. The green boxplots represent the performance achieved with our balancing method IMBALMED across the folds of the cross-validation, whilst the blue boxplots depict the performance from the unbalanced approach, also across the 10 folds. A green and blue line connects the average performance for each classifier from the balanced and unbalanced methods, respectively. The boxplots are organized in descending order, starting with the classifier that achieved the highest average G-mean across the 10 folds for the balanced dataset method.
Across all conducted comparisons, our method consistently demonstrated higher average performance than the unbalanced approach. Notably, the 12-month early detection task (Figure~\ref{fig:12progn}) showed the greatest difference in performance between the balanced and unbalanced methods, indicating a significant improvement. This is attributed to the more pronounced imbalance in this task's dataset, where the majority class CN+MCI constitutes 91.34\% of the recruited patients.

\begin{table}[htbp]
\centering
\begin{subtable}[b]{\textwidth}
    \centering
    \small 
    \resizebox{.8\textwidth}{!}{
    \begin{tabular}{cc|cc|cc|}
    \cline{3-6}
    &      & \multicolumn{4}{c|}{\textbf{Diagnostic tasks}}  \\ \cline{3-6} 
    &      & \multicolumn{2}{c|}{\textbf{Binary}} & \multicolumn{2}{c|}{\textbf{Ternary}}                  \\ \cline{3-6} 
     &      & \multicolumn{1}{c|}{\textbf{W-T-L}} & \multicolumn{1}{c|}{\textbf{G-mean}} & \multicolumn{1}{c|}{\textbf{W-T-L}} & \multicolumn{1}{c|}{\textbf{G-mean}} \\ \hline
    \multicolumn{1}{|c|}{\multirow{12}{*}{\rotatebox{90}{Algorithm}}} & IMBALMED & \multicolumn{1}{c|}{-} & \cellcolor{lightgray}\textbf{96.98 ± 2.00} & \multicolumn{1}{c|}{-} & \cellcolor{lightgray}\textbf{72.36 ± 3.90} \\ \cline{2-6} 
    \multicolumn{1}{|c|}{} & ARF & \multicolumn{1}{c|}{90-0-10 (W*)} & 95.19 ± 2.15 & \multicolumn{1}{c|}{60-0-40 (W)} & 70.99 ± 5.05 \\ \cline{2-6}
    \multicolumn{1}{|c|}{}  & ARFR & \multicolumn{1}{c|}{90-0-10 (W*)} & 95.23 ± 2.12 & \multicolumn{1}{c|}{70-0-30 (W)} & 71.40 ± 3.94 \\ \cline{2-6}
    \multicolumn{1}{|c|}{}  & CALMID & \multicolumn{1}{c|}{70-0-30 (W*)} & 95.27 ± 2.41 & \multicolumn{1}{c|}{100-0-0 (W*)} & 64.68 ± 3.47 \\ \cline{2-6}
    \multicolumn{1}{|c|}{} & LB & \multicolumn{1}{c|}{80-0-20 (W*)} & 95.31 ± 2.29 & \multicolumn{1}{c|}{90-0-10 (W*)} & 67.40 ± 4.55 \\ \cline{2-6}
    \multicolumn{1}{|c|}{} & OADA & \multicolumn{1}{c|}{80-0-20 (W*)} & 95.01 ± 2.24 & \multicolumn{1}{c|}{100-0-0 (W*)} & 65.59 ± 4.35 \\ \cline{2-6}
    \multicolumn{1}{|c|}{} & OBA & \multicolumn{1}{c|}{90-0-10 (W*)} & 94.80 ± 2.76 & \multicolumn{1}{c|}{100-0-0 (W*)} & 66.89 ± 2.58 \\ \cline{2-6}
    \multicolumn{1}{|c|}{} & OOB & \multicolumn{1}{c|}{90-0-10 (W*)} & 94.71 ± 3.07 & \multicolumn{1}{c|}{100-0-0 (W*)} & 65.54 ± 2.87 \\ \cline{2-6}
    \multicolumn{1}{|c|}{} & ROSE & \multicolumn{1}{c|}{80-0-20 (W*)} & 95.31 ± 2.29 & \multicolumn{1}{c|}{80-0-20 (W*)} & 69.93 ± 4.74 \\ \cline{2-6}
    \multicolumn{1}{|c|}{} & UOB & \multicolumn{1}{c|}{100-0-0 (W*)} & 94.02 ± 2.69 & \multicolumn{1}{c|}{100-0-0 (W*)} & 66.64 ± 3.10 \\ \hline
    \end{tabular}}
    \caption{Diagnostic tasks: binary and ternary.}
    \label{tab:diagnostic_tasks}
\end{subtable}

\vspace{0.5cm}

\begin{subtable}[b]{\textwidth}
    \centering
    \small
    \resizebox{.8\textwidth}{!}{
    \begin{tabular}{cc|cc|cc|}
    \cline{3-6}
     &      & \multicolumn{4}{c|}{\textbf{Early detection tasks}}                           \\ \cline{3-6} 
     &      & \multicolumn{2}{c|}{\textbf{12 months}} & \multicolumn{2}{c|}{\textbf{24 months}} \\ \cline{3-6} 
     &      & \multicolumn{1}{c|}{\textbf{W-T-L}} & \multicolumn{1}{c|}{\textbf{G-mean}} & \multicolumn{1}{c|}{\textbf{W-T-L}} & \multicolumn{1}{c|}{\textbf{G-mean}} \\ \hline
    \multicolumn{1}{|c|}{\multirow{12}{*}{\rotatebox{90}{Algorithm}}} & IMBALMED & \multicolumn{1}{c|}{-} & \cellcolor{lightgray}\textbf{76.53 ± 4.18} & \multicolumn{1}{c|}{-} & \textbf{82.68 ± 2.58} \\ \cline{2-6} 
    \multicolumn{1}{|c|}{} & ARF & \multicolumn{1}{c|}{90-0-10 (W*)} & 49.42 ± 13.86 & \multicolumn{1}{c|}{70-0-30 (W*)} & 76.19 ± 8.17 \\ \cline{2-6}
    \multicolumn{1}{|c|}{}  & ARFR & \multicolumn{1}{c|}{80-0-20 (W*)} & 65.75 ± 12.62 & \multicolumn{1}{c|}{30-0-70 (L)} & 82.97 ± 4.26 \\ \cline{2-6}
    \multicolumn{1}{|c|}{}  & CALMID & \multicolumn{1}{c|}{100-0-0 (W*)} & 75.09 ± 7.75 & \multicolumn{1}{c|}{70-0-30 (W)} & 71.46 ± 24.44 \\ \cline{2-6}
    \multicolumn{1}{|c|}{} & LB & \multicolumn{1}{c|}{100-0-0 (W*)} & 75.09 ± 7.75 & \multicolumn{1}{c|}{70-0-30 (W*)} & 74.68 ± 9.54 \\ \cline{2-6}
    \multicolumn{1}{|c|}{} & OADA & \multicolumn{1}{c|}{90-0-10 (W*)} & 68.49 ± 7.36 & \multicolumn{1}{c|}{60-0-40 (W*)} & 78.73 ± 5.08 \\ \cline{2-6}
    \multicolumn{1}{|c|}{} & OBA & \multicolumn{1}{c|}{100-0-0 (W*)} & 11.52 ± 23.06 & \multicolumn{1}{c|}{70-0-30 (W)} & 72.78 ± 24.69 \\ \cline{2-6}
    \multicolumn{1}{|c|}{} & OOB & \multicolumn{1}{c|}{70-0-30 (W*)} & 71.35 ± 6.87 & \multicolumn{1}{c|}{40-0-60 (L)} & 82.02 ± 4.71 \\ \cline{2-6}
    \multicolumn{1}{|c|}{} & ROSE & \multicolumn{1}{c|}{80-0-20 (W*)} & 60.04 ± 15.96 & \multicolumn{1}{c|}{80-0-20 (W*)} & 76.08 ± 10.27 \\ \cline{2-6}
    \multicolumn{1}{|c|}{} & UOB & \multicolumn{1}{c|}{60-0-40 (W)} & 74.57 ± 4.00 & \multicolumn{1}{c|}{60-0-40 (W)} & 81.32 ± 5.31 \\ \hline
    \end{tabular}}
    \caption{Early detection tasks: 12 and 24 months.}
    \label{tab:prognostic_tasks_12_24}
\end{subtable}

\vspace{0.5cm}

\begin{subtable}[b]{\textwidth}
    \centering
    \small
    \resizebox{.8\textwidth}{!}{
    \begin{tabular}{cc|cc|cc|}
    \cline{3-6}
     &      & \multicolumn{4}{c|}{\textbf{Early detection tasks}} \\ \cline{3-6} 
     &      & \multicolumn{2}{c|}{\textbf{36 months}} & \multicolumn{2}{c|}{\textbf{48 months}}                \\ \cline{3-6} 
     &      & \multicolumn{1}{c|}{\textbf{W-T-L}} & \multicolumn{1}{c|}{\textbf{G-mean}} & \multicolumn{1}{c|}{\textbf{W-T-L}} & \multicolumn{1}{c|}{\textbf{G-mean}} \\ \hline
    \multicolumn{1}{|c|}{\multirow{12}{*}{\rotatebox{90}{Algorithm}}} & IMBALMED & \multicolumn{1}{c|}{-} & \cellcolor{lightgray}\textbf{81.44 ± 3.07} & \multicolumn{1}{c|}{-} & \cellcolor{lightgray}\textbf{85.66 ± 2.43} \\ \cline{2-6} 
    \multicolumn{1}{|c|}{} & ARF & \multicolumn{1}{c|}{100-0-0 (W*)} & 72.40 ± 8.18 & \multicolumn{1}{c|}{90-0-10 (W*)} & 81.07 ± 3.29 \\ \cline{2-6}
    \multicolumn{1}{|c|}{}  & ARFR & \multicolumn{1}{c|}{60-0-40 (W*)} & 75.36 ± 8.36 & \multicolumn{1}{c|}{80-0-20 (W)} & 84.19 ± 3.99 \\ \cline{2-6}
    \multicolumn{1}{|c|}{}  & CALMID & \multicolumn{1}{c|}{80-0-20 (W*)} & 64.57 ± 25.23 & \multicolumn{1}{c|}{60-0-40 (W)} & 83.10 ± 6.24 \\ \cline{2-6}
    \multicolumn{1}{|c|}{} & LB & \multicolumn{1}{c|}{100-0-0 (W*)} & 65.76 ± 12.01 & \multicolumn{1}{c|}{60-0-40 (W)} & 83.10 ± 6.24 \\ \cline{2-6}
    \multicolumn{1}{|c|}{} & OADA & \multicolumn{1}{c|}{60-0-40 (W*)} & 78.14 ± 6.17 & \multicolumn{1}{c|}{90-0-10 (W*)} & 81.83 ± 4.61 \\ \cline{2-6}
    \multicolumn{1}{|c|}{} & OBA & \multicolumn{1}{c|}{50-0-50 (X)} & 80.04 ± 6.74 & \multicolumn{1}{c|}{60-0-40 (W)} & 84.28 ± 4.28 \\ \cline{2-6}
    \multicolumn{1}{|c|}{} & OOB & \multicolumn{1}{c|}{40-0-60 (L)} & 79.85 ± 6.95 & \multicolumn{1}{c|}{60-0-40 (W)} & 83.56 ± 5.16 \\ \cline{2-6}
    \multicolumn{1}{|c|}{} & ROSE & \multicolumn{1}{c|}{80-0-20 (W*)} & 77.12 ± 6.35 & \multicolumn{1}{c|}{90-0-10 (W*)} & 80.26 ± 4.18 \\ \cline{2-6}
    \multicolumn{1}{|c|}{} & UOB & \multicolumn{1}{c|}{70-10-20 (W)} & 78.17 ± 6.74 & \multicolumn{1}{c|}{70-0-30 (W*)} & 82.22 ± 4.51 \\ \hline
    \end{tabular}}

    \caption{Early detection tasks: 36 and 48 months.}
    \label{tab:prognostic_tasks_36_48}
\end{subtable}

\caption{Balancing data competitors on diagnostic (table a) and early detection tasks (tables b and c): W for win rates exceeding tie and loss rates, L for highest loss rates, X for equal win and loss rates, and * for statistically significant outcomes. Tasks with a win rate above half of the competitors are highlighted in bold, while our superior performances are shaded gray.}
\label{tab:diagnostic_prognostic_tasks}
\end{table}

In our comparative analysis against nine cutting-edge algorithms for imbalanced data, Table~\ref{tab:diagnostic_prognostic_tasks} shows the average G-mean across the cross-validation folds for each algorithm. As previously mentioned in Section~\ref{subsec:competitor}, for this comparative analysis we evaluated the performance of state-of-the-art algorithms against the best-performing classifier with our proposed balancing method IMBALMED, which was selected based on its performance on an independent validation set. Our top-performing classifiers in the validation set were: AB for the binary diagnostic task, LDA for the ternary diagnostic, LS for the 12-month early detection task, LR for the 24- and 48-month early detection tasks, and HGB for the 36-month early detection task. Their performances on the test set are reported in the IMBALMED rows for each task.
The W-T-L columns in Table~\ref{tab:diagnostic_prognostic_tasks} report the win, tie, and loss rates, respectively. For each competitor, these columns also include a notation in brackets: W if the win rate exceeds both the tie and loss rates; L if the loss rate dominates; X when the win rate and loss rate are equal. We highlight in bold the tasks where our performance demonstrate a higher win rate against more than half of the competitors. Tasks where IMBALMED's performance is superior to the competitors are shaded with a gray background. Notably, experiments indicating statistical significance with win/tie/loss rates are highlighted with a * symbol.

For the binary diagnostic task (Table~\ref{tab:diagnostic_tasks}), our method showed superior performance in all nine experiments, with statistically significant win rates across the board. In the ternary task (Table~\ref{tab:diagnostic_tasks}), our balancing method outperformed all competitors, with seven out nine of these having statistical significance win rates that exceeded both tie and loss rates. 
Both binary and ternary tasks had no experiments where the loss rate was higher than both win and tie rates. Notably, in the binary task, our method showed the lowest standard deviation, highlighting its robustness of our method. Although in the ternary task our standard deviation wasn’t the lowest, we still achieved a 1.34\% performance improvement over the top competitor, the ARFR algorithm.

In the 12-month early detection task (Table~\ref{tab:prognostic_tasks_12_24}), IMBALMED surpassed all competitors, with a win rate higher than both tie and loss rates, and eight of these showing statistical significance. In this 12-month task, IMBALMED had the highest performance, improving by 1.92\% over the best competitors CALMID and LB.
For the 24-month early detection task (Table~\ref{tab:prognostic_tasks_12_24}), our method outperformed eight out of nine competitors, with seven of these showing higher win rates than tie and loss rates, and four of these had statistical significance. The ARFR classifier achieved the highest performance, improving ours by 0.35\%; however, it is notable that the standard deviation of our method is 39.44\% lower than that obtained by the ARFR algorithm. Additionally, in comparison with the OOB algorithm, our method demonstrates a loss rate higher than both the win rate and the tie rate. This indicates that in more than half of the 10-folds, OOB's performance is superior, but the overall average across the 10-folds is higher with our method, allowing a performance increase of 0.80\% with IMBALMED and a standard deviation 45.22\% lower.
In the 36-month task (Table~\ref{tab:prognostic_tasks_36_48}), IMBALMED was superior to all competitors, with seven of them having a higher win rate than both tie and loss rates, and six of these showing statistical significance. It is noteworthy that the OBA algorithm demonstrates equal win and loss rates, with a performance 1.75\% lower than ours. Additionally, the OOB algorithm shows a higher win rate compared to our method, but with a performance 1.99\% lower and a standard deviation 55.83\% higher, underscoring the robustness of our method.
In the 48-month early detection task (Table~\ref{tab:prognostic_tasks_36_48}), our method excelled across all competitors, with every experiment showing a higher win rate than both tie and loss rates, four of which had statistical significance. In this 48-month task, IMBALMED had the highest performance, improving by 1.64\% over the best competitor (OBA).
It is important to emphasize that in the clinically significant 48-month task, our IMBALMED method achieved the highest performance compared to other competitors. The 48-month early detection task is particularly crucial because it allows for a longer-term forecast of AD progression, providing critical insights for early intervention and treatment planning.  Shorter-term predictions at 12, 24, or 36 months, while useful, do not offer the same strategic advantage.

Overall, IMBALMED demonstrated superior performance across most tasks and timeframes. By addressing class imbalance and enhancing model robustness, IMBALMED offers a powerful tool for clinicians to forecast disease progression, enabling timely interventions. The outperformance of IMBALMED, especially in the critical 48-month early detection task, underscores its potential to significantly advance the field of AD early detection and care.

\section{Conclusions}\label{sec:Conclusions}
The rising prevalence of AD, the limited sensitivity of traditional diagnostic methods, and the societal and economic burdens it imposes underline the critical need for advanced diagnostic and early detection methodologies. In this study we introduced a novel methodology for addressing data imbalance, that leverages the strengths of multimodal ensemble learning while incorporating multi-level data fusion. For each of the four selected tabular modalities from the ADNI database~\cite{ADNI}, we train a series of classifiers on varied class distributions followed by a fusion strategy that integrates the different modalities. Our framework was evaluated on two diagnostic tasks (binary and ternary) and four binary early detection tasks (at 12, 24, 36, and 48 months) and it was compared with the standard unbalanced dataset approach and with state-of-the-art imbalanced data algorithms. 
Through our proposed multi-level data fusion strategy, our method IMBALMED has proven superior to unbalanced approaches, enhancing diversity and effectively boosting performance.
Comparative analysis with state-of-the-art algorithms revealed that our approach enhances performance in all experiments on binary and ternary diagnostic tasks, 12-, 36-, and 48-month early detection tasks experiments, and eight out of nine experiments on the 24-month early detection task.
Additionally, we observed statistically significant win rates in 100\% of binary diagnostic tasks, 77.78\% of ternary diagnostic tasks, and 88.89\%, 44.44\%, 66.67\%, and 44.44\% of 12-, 24-, 36-, and 48-month tasks, respectively, all without any statistically significant loss rates.

A limitation of our study is that all patients were sourced from a single dataset, the ADNI. Whilst this was acceptable for our research, a clinical implementation of our model would benefit from additional datasets, even of different diseases, to capture greater feature variability and to study performance variation. Furthermore, although unimodal analyses have shown that all modalities are nearly equally informative for early detection tasks, the diagnostic task reveals a greater influence of the Assessment modality. Therefore, conducting an ablation study on modalities would be significant, as it could further enhance performance.

\section*{CRediT authorship contribution statement}
\textbf{Arianna Francesconi:} Writing – original draft, Writing – review \& editing, Visualization, Validation, Software, Methodology, Investigation, Formal analysis, Data curation, Conceptualization. 
\textbf{Lazzaro di Biase:} Conceptualization, Supervision.
\textbf{Donato Cappetta:} Funding acquisition.
\textbf{Fabio Rebecchi:} Funding acquisition.
\textbf{Paolo Soda:} Writing – review \& editing, Supervision, Project administration, Methodology, Funding acquisition, Conceptualization.
\textbf{Rosa Sicilia:} Writing – review \& editing, Validation, Supervision, Software, Project administration, Methodology, Visualization, Formal analysis, Conceptualization. 
\textbf{Valerio Guarrasi:} Writing – original draft, Writing – review \& editing, Validation, Supervision, Software, Project administration, Methodology, Visualization, Formal analysis, Conceptualization.

\section*{Declaration of competing interest}
The authors declare that they have no known competing financial interests or personal relationships that could have appeared to influence the work reported in this paper.

\section*{Acknowledgments}
Arianna Francesconi is a Ph.D. student enrolled in the National Ph.D. in Artificial Intelligence, XXXIX cycle, course on Health and Life Sciences, organized by Università Campus Bio-Medico di Roma. This work was partially founded by: i) PNRR – DM 117/2023; ii) Eustema S.p.A.; iii) PNRR MUR, Italy project PE0000013 - FAIR.
Data collection and sharing for ADNI was funded by the Alzheimer’s Disease Neuroimaging Initiative (ADNI) (National Institutes of Health Grant U01 AG024904) and DOD ADNI (Department of Defense award number W81XWH-12-2-0012). ADNI is funded by the National Institute on Aging (National Institutes of Health Grant U19AG024904) and the grantee organization is the Northern California Institute for Research and Education. In the past, ADNI has also received funding from the National Institute of Biomedical Imaging and Bioengineering, the Canadian Institutes of Health Research, and private sector contributions through the Foundation for the National Institutes of Health (FNIH) including generous contributions from the following: AbbVie, Alzheimer's Association; Alzheimer's Drug Discovery Foundation; Araclon Biotech; BioClinica, Inc.; Biogen; Bristol-Myers Squibb Company; CereSpir, Inc.; Cogstate; Eisai Inc.; Elan Pharmaceuticals, Inc.; Eli Lilly and Company; EuroImmun; F. Hoffmann-La Roche Ltd and its affiliated company Genentech, Inc.; Fujirebio; GE Healthcare; IXICO Ltd.; Janssen Alzheimer Immunotherapy Research \& Development, LLC.; Johnson \& Johnson Pharmaceutical Research \& Development LLC.; Lumosity; Lundbeck; Merck \& Co., Inc.; Meso Scale Diagnostics, LLC.; NeuroRx Research; Neurotrack Technologies; Novartis Pharmaceuticals Corporation; Pfizer Inc.; Piramal Imaging; Servier; Takeda Pharmaceutical Company; and Transition Therapeutics.

\bibliographystyle{elsarticle-num-names} 
\bibliography{bibliography.bib}

\newpage
\appendix

\section{Features selected}\label{app:features}

\begin{table}[ht]
\renewcommand{\arraystretch}{1.5} 
\begin{adjustbox}{width=\textwidth}
\begin{tabular}{|l|l|l|}
\hline
\multicolumn{1}{|c|}{\textbf{Modality}} & \multicolumn{1}{c|}{\textbf{Diagnostic Tasks}} & \multicolumn{1}{c|}{\textbf{Early Detection Tasks}}                           \\ \hline
Assessment & \begin{tabular}[c]{@{}l@{}}FAQ\_bl, FAQSOURCE, FAQFINAN, FAQFORM, FAQSHOP, FAQGAME,\\  FAQBEVG, FAQMEAL, FAQEVENT, FAQTV, FAQREM, FAQTRAVL, \\ HMONSET, HMSTEPWS, HMSOMATC, HMEMOTIO, HMHYPERT, \\ HMSTROKE, HMNEURSM, HMNEURSG, HMSCORE, GDSATIS, \\ GDDROP, GDEMPTY, GDBORED, GDSPIRIT, GDAFRAID, GDHAPPY, \\ GDHELP, GDHOME, GDMEMORY, GDALIVE, GDWORTH, GDENERGY, \\ GDHOPE, GDBETTER, GDTOTAL\end{tabular} & \begin{tabular}[c]{@{}l@{}}FAQ\_bl, FAQSOURCE, FAQFINAN, FAQFORM, FAQSHOP, FAQGAME, \\ FAQBEVG, FAQMEAL, FAQEVENT, FAQTV, FAQREM, FAQTRAVL, \\ HMONSET, HMSTEPWS, HMSOMATC, HMEMOTIO, HMHYPERT, \\ HMSTROKE, HMNEURSM, HMNEURSG, HMSCORE, GDSATIS, GDDROP, \\ GDEMPTY, GDBORED, GDSPIRIT, GDAFRAID, GDHAPPY, GDHELP, \\ GDHOME, GDMEMORY, GDALIVE, GDWORTH, GDENERGY, GDHOPE, \\ GDBETTER, GDTOTAL, PHC\_Diagnosis, PHC\_MEM, PHC\_EXF, PHC\_LAN, \\ WORD1, WORD2, WORD3, MMWATCH, MMPENCIL, MMREPEAT, \\ MMHAND, MMFOLD, MMONFLR, MMREAD, MMWRITE, \\ MMDRAW, MMSCORE, CLOCKCIRC, CLOCKSYM, CLOCKNUM, \\ CLOCKHAND, CLOCKTIME, CLOCKSCOR, COPYCIRC, COPYSYM, \\ COPYNUM, COPYHAND, COPYTIME, COPYSCOR, AVTOT1, AVERR1, \\ AVTOT2, AVERR2, AVTOT3, AVERR3, AVTOT4, AVERR4, AVTOT5, \\ AVERR5, AVTOT6, AVERR6, AVTOTB, AVERRB, CATANIMSC, \\ CATANPERS, CATANINTR, TRAASCOR, TRAAERRCOM, TRAAERROM, \\ TRABSCOR, TRABERRCOM, TRABERROM, AVDEL30MIN, AVDELERR1, \\ AVDELTOT, AVDELERR2, ADNI\_MEM, ADNI\_EF, ADNI\_LAN, \\ ADNI\_VS, ADNI\_EF2, FAQ\_bl.1, MMSE\_bl, CDRSB\_bl, RAVLT\_immediate\_bl, \\ RAVLT\_learning\_bl, RAVLT\_forgetting\_bl, RAVLT\_perc\_forgetting\_bl, \\ LDELTOTAL\_BL, TRABSCOR\_bl, mPACCdigit\_bl, mPACCtrailsB\_bl, DX\_bl\end{tabular} \\ \hline
Biospecimen & APOE4 & APOE4, TAU\_ADNIMERGE, PTAU\_ADNIMERGE, DX\_bl                                    \\ \hline
Image Analysis & \begin{tabular}[c]{@{}l@{}}FDG\_bl, FDG, Ventricles\_bl, Hippocampus\_bl, WholeBrain\_bl, \\ Entorhinal\_bl, Fusiform\_bl, MidTemp\_bl, ICV\_bl\end{tabular} & \begin{tabular}[c]{@{}l@{}}FDG\_bl, FDG, Ventricles\_bl, Hippocampus\_bl, WholeBrain\_bl, \\ Entorhinal\_bl, Fusiform\_bl, MidTemp\_bl, ICV\_bl, DX\_bl\end{tabular}                  \\ \hline
Subject Characteristics & AGE, PTGENDER, PTEDUCAT, PTETHCAT, PTRACCAT, PTMARRY & AGE, PTGENDER, PTEDUCAT, PTETHCAT, PTRACCAT, PTMARRY, DX\_bl \\ \hline
\end{tabular}
\end{adjustbox}
\caption{Detailed list of selected features by modality and task.}
\label{tab:features_selected}
\end{table}


\section{Unimodal Fusion Performance}\label{app:unimodal}

\begin{table}[ht]
\begin{adjustbox}{width=\textwidth}
\begin{tabular}{|l|l|l|l|l|l|l|}
\hline
\multicolumn{1}{|c|}{\textbf{Modality}} & \multicolumn{1}{c|}{\textbf{\begin{tabular}[c]{@{}c@{}}Binary\\ Diagnostic\end{tabular}}} & \multicolumn{1}{c|}{\textbf{\begin{tabular}[c]{@{}c@{}}Ternary \\ Diagnostic\end{tabular}}} & \multicolumn{1}{c|}{\textbf{\begin{tabular}[c]{@{}c@{}}12-month\\ Early detection\end{tabular}}} & \multicolumn{1}{c|}{\textbf{\begin{tabular}[c]{@{}c@{}}24-month\\ Early detection\end{tabular}}} & \multicolumn{1}{c|}{\textbf{\begin{tabular}[c]{@{}c@{}}36-month\\ Early detection\end{tabular}}} & \multicolumn{1}{c|}{\textbf{\begin{tabular}[c]{@{}c@{}}48-month \\ Early detection\end{tabular}}} \\ \hline
Assessment & 95.42 ± 0.72 & 71.90 ± 2.40 & 78.50 ± 3.33 & 83.71 ± 2.25 & 81.58 ± 1.61 & 84.78 ± 3.17 \\ \hline
Biospecimen     & 66.85 ± 4.69 & 25.98 ± 9.30 & 71.21 ± 1.42 & 78.90 ± 1.19 & 75.15 ± 1.98 & 81.41 ± 1.25 \\ \hline
Image Analysis  & 85.52 ± 1.29 & 53.35 ± 2.19 & 74.77 ± 2.01 & 81.82 ± 1.11 & 78.92 ± 1.75 & 83.14 ± 1.06 \\ \hline
\begin{tabular}[c]{@{}l@{}}Subject Characteristics\end{tabular} & 59.40 ± 3.47  & 38.00 ± 5.50 & 69.41 ± 1.82 & 78.01 ± 2.31 & 74.64 ± 2.34 & 78.65 ± 3.97 \\ \hline
\end{tabular}
\end{adjustbox}
\caption{Performance comparison of unimodal modalities for various tasks. The table shows the effectiveness of different modalities in diagnostic and early detection tasks, including performance differences in binary and ternary classification tasks and across different time points.}
\label{tab:unimodal_results}
\end{table}

\end{document}